%% file: example_paper.tex
\documentclass{article}

\usepackage{courier}
\usepackage{microtype}
\usepackage{graphicx}
\usepackage{mathtools}
\usepackage{pifont}
\usepackage{algorithmic}
\usepackage{subfigure}
\usepackage{makecell}
\usepackage{multirow,multicol}
\usepackage{booktabs} 

\usepackage{hyperref}

\newcommand{\logopic}{%
   \raisebox{-0.5ex}{
      \includegraphics[height=2.5ex]{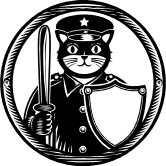}
  }%
}

\usepackage[accepted]{icml2024}

\usepackage{amsmath}
\usepackage{amssymb}
\usepackage{mathtools}
\usepackage{amsthm}
\usepackage{colortbl}

\usepackage[capitalize,noabbrev]{cleveref}
\theoremstyle{plain}

\theoremstyle{definition}

\theoremstyle{remark}

\usepackage[textsize=tiny]{todonotes}

\definecolor{myblue}{RGB}{0, 102, 204} 

\icmltitlerunning{EraseAnything}

\begin{document}

\twocolumn[
\icmltitle{\logopic EraseAnything: Enabling Concept Erasure in Rectified Flow Transformers}





\begin{icmlauthorlist}
\icmlauthor{Daiheng Gao}{aaa,hhh}
\icmlauthor{Shilin Lu}{bbb}
\icmlauthor{Shaw Walters}{hhh}
\icmlauthor{Wenbo Zhou}{aaa}
\icmlauthor{Jiaming Chu}{ccc}
\icmlauthor{Jie Zhang}{ddd}
\icmlauthor{Bang Zhang}{eee}
\icmlauthor{Mengxi Jia}{ggg}
\icmlauthor{Jian Zhao}{ggg}
\icmlauthor{Zhaoxin Fan}{fff}
\icmlauthor{Weiming Zhang}{aaa}
\end{icmlauthorlist}


\begin{icmlauthorlist}
{$^1$USTC}
{$^2$Eliza Labs}
{$^3$NTU}
{$^4$BUPT}
{$^5$A*STAR}
{$^6$Alibaba Tongyi lab}
{$^7$TeleAI}
{$^8$Beihang University}
\end{icmlauthorlist} 


\icmlkeywords{Machine Learning, ICML}

\vskip 0.3in

]

\newcommand{\eg}{\textit{e.g.}}



\definecolor{SigmaColor}{rgb}{0.98,0.45,0.0}
\definecolor{AlphaColor}{rgb}{0,0,0.8}
\definecolor{BetaColor}{rgb}{0.8,0,0.8}

\begin{abstract}
Removing unwanted concepts from large-scale text-to-image (T2I) diffusion models while maintaining their overall generative quality remains an open challenge. This difficulty is especially pronounced in emerging paradigms, such as Stable Diffusion (SD) v3 and Flux, which incorporate flow matching and transformer-based architectures. These advancements limit the transferability of existing concept-erasure techniques that were originally designed for the previous T2I paradigm (\textit{e.g.}, SD v1.4).  In this work, we introduce \logopic \textbf{EraseAnything}, the first method specifically developed to address concept erasure within the latest flow-based T2I framework. We formulate concept erasure as a bi-level optimization problem, employing LoRA-based parameter tuning and an attention map regularizer to selectively suppress undesirable activations. Furthermore, we propose a self-contrastive learning strategy to ensure that removing unwanted concepts does not inadvertently harm performance on unrelated ones. Experimental results demonstrate that EraseAnything successfully fills the research gap left by earlier methods in this new T2I paradigm, achieving state-of-the-art performance across a wide range of concept erasure tasks. 
\end{abstract}

\input{sec/introduction}
\input{sec/related_work}
\input{sec/motivation}
\input{sec/method}
\input{sec/experiment}
\input{sec/conclusion}
\bibliography{example_paper}
\bibliographystyle{icml2024}

\newpage
\appendix
\onecolumn



\section{Flux Architecture}
\label{sec:app_1}

In our research, we have chosen Flux [dev] as our baseline model due to its reputation as the most performant within the open-source Flux series~\footnote{https://blackforestlabs.ai/announcing-black-forest-labs/}. As highlighted in \cref{sec:sec3}, Flux's architecture significantly diverges from that of SD v1.5, which has been the predominant baseline for contemporary concept erasure techniques.

\begin{figure*}[bh]
\centering
\includegraphics[width=0.8\textwidth]{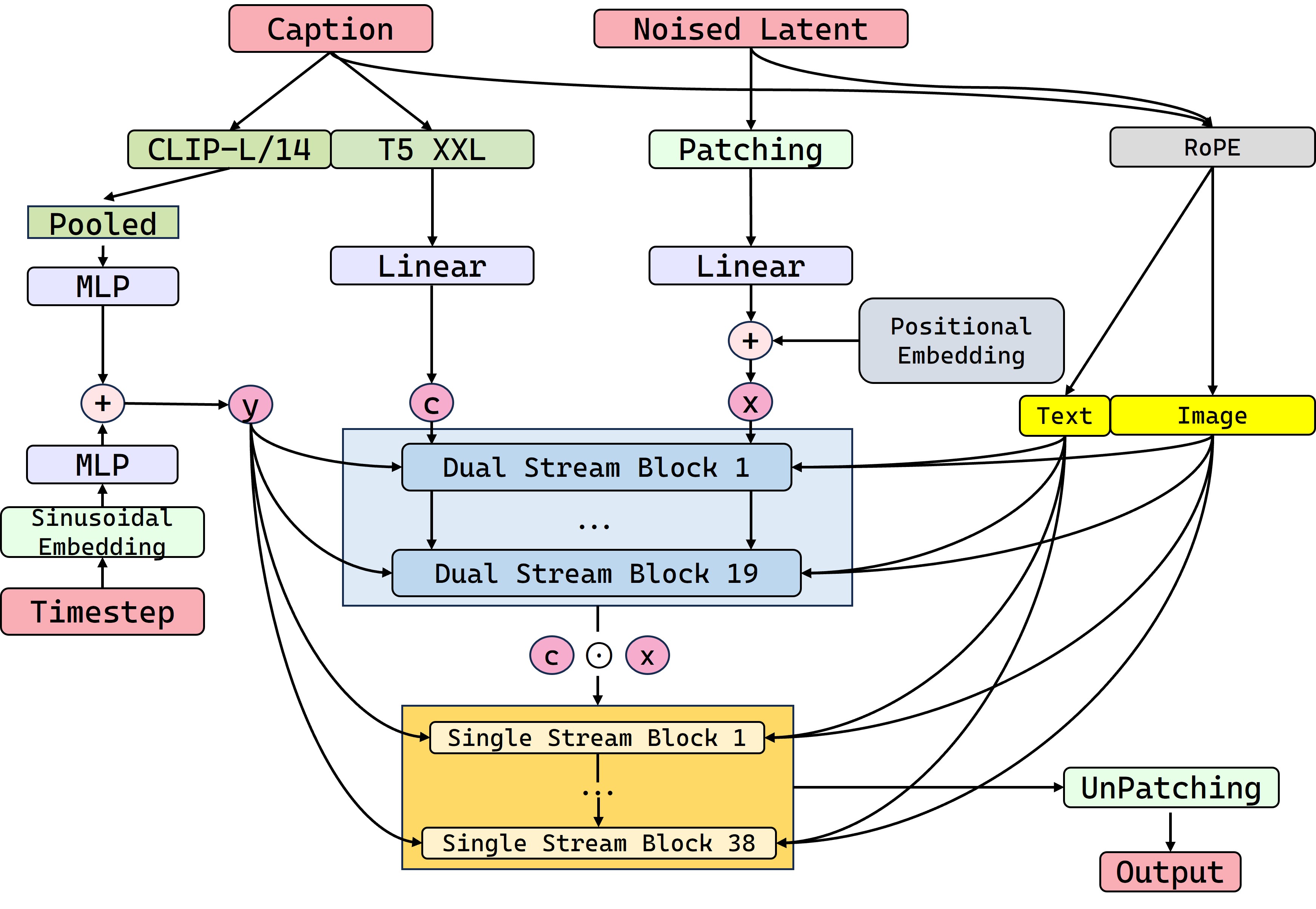}
\caption{\textbf{Model architecture of Flux [dev]}. Flux [dev] use frozen CLIP-L 14 and T5-XXL as text encoders for conditioned caption
feature extraction. The coarsed CLIP embedding concatenated with
timestep embedding $y$ are used to modulation mechanism. The
fine-grained T5 $c$ concatenated with image latents $x$ are input
to a stacked of double stream blocks and single stream blocks to
predict output in the VAE encoded latent space. Concatenation is indicated by $\odot$.}
\label{fig:sup_1}
\vspace{-0.1in}
\end{figure*}

\begin{figure*}
\centering
\includegraphics[width=0.8\textwidth]{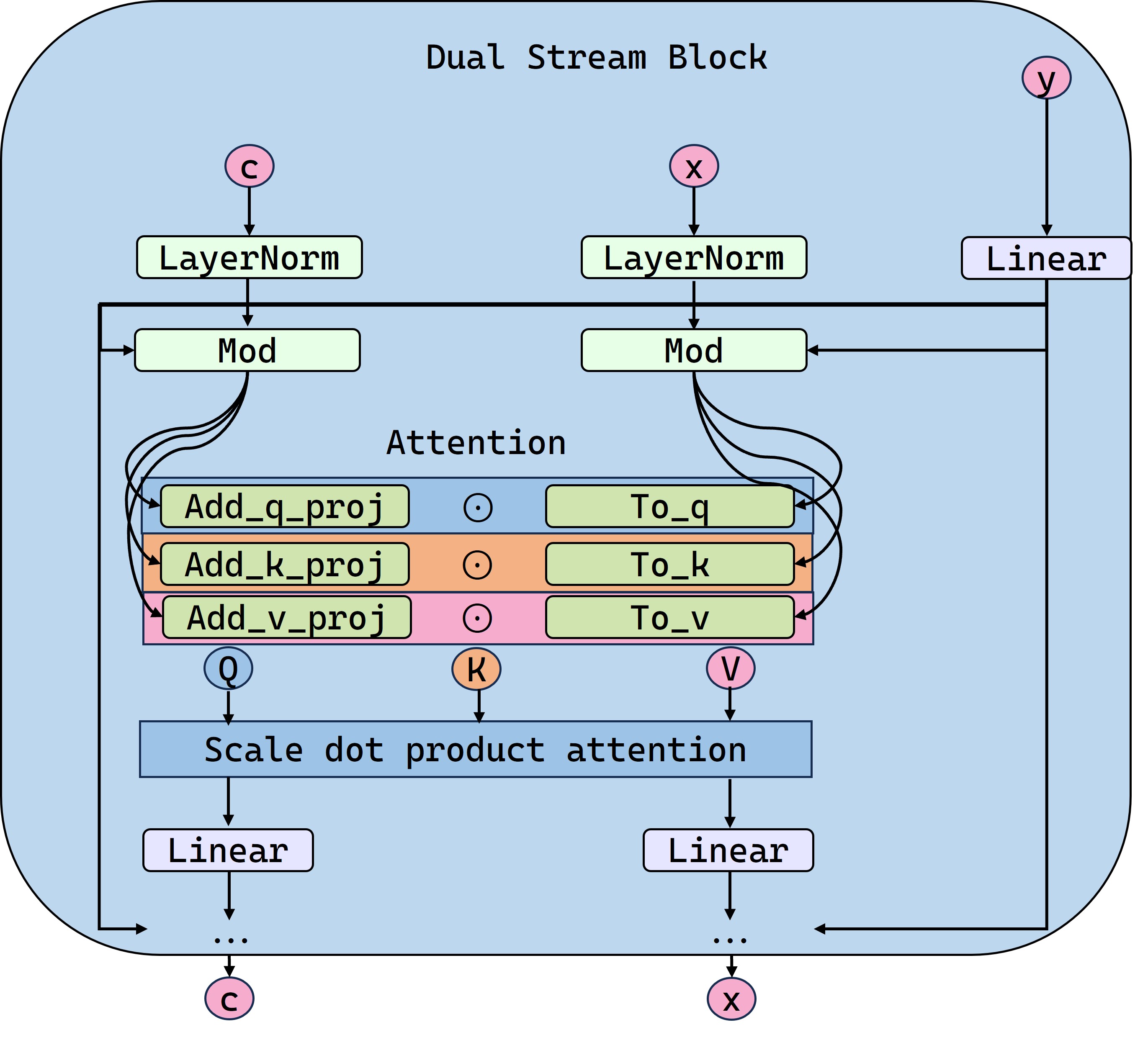}
\caption{\textbf{Dual stream block}. In Flux, the semantic correlation is established in the dual stream block, which established an implicit relationshio between text and image. Noteworthy thing is that the explicit cross attention module that prevails among SD v1.5 is not existed in Flux.}
\label{fig:sup_2}
\vspace{-0.2in}
\end{figure*}

\begin{figure*}
\centering
\includegraphics[width=0.8\textwidth]{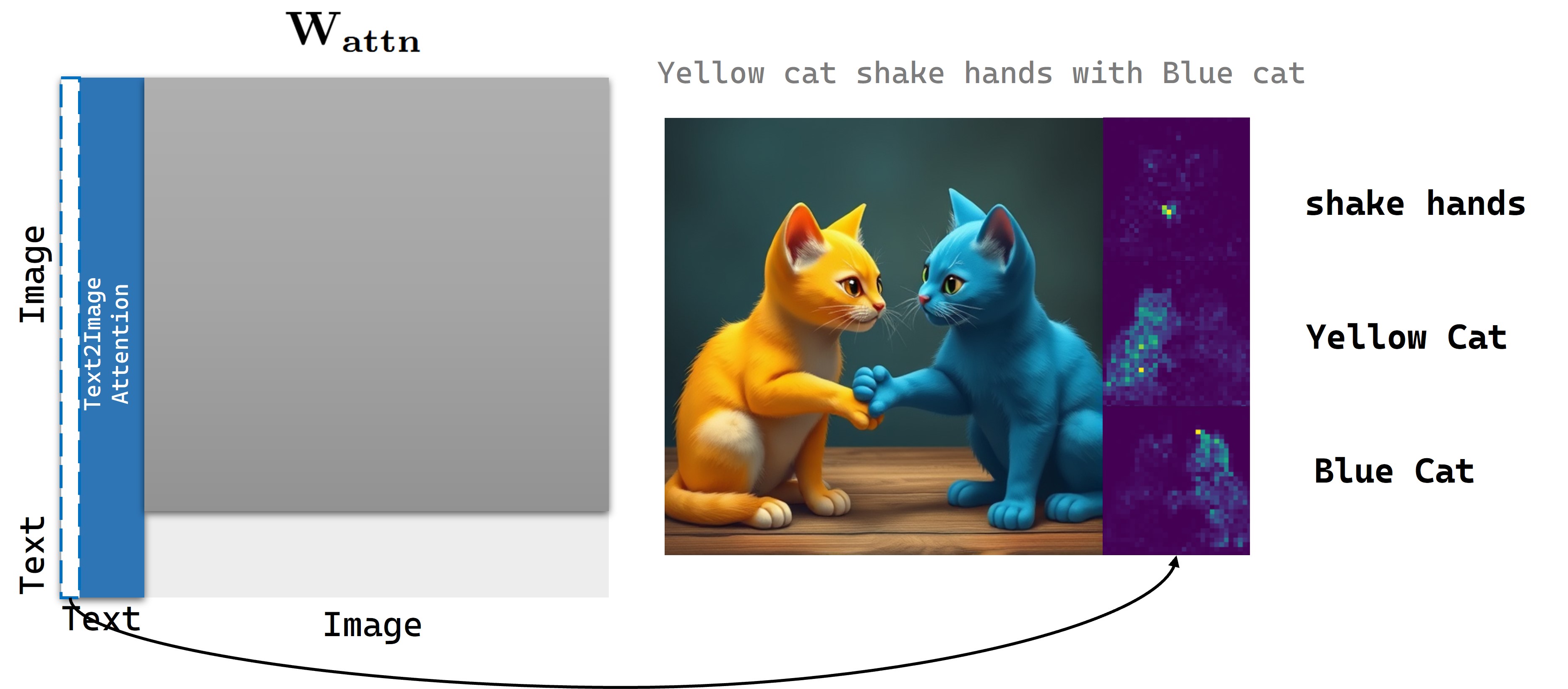}
\caption{\textbf{Attention map extraction}. The correlation between specific words and their corresponding heatmaps can be discerned within the matrix $\mathbf{W_{attn}}$, particularly within the columns (\textbf{white} bar adorned with a \textbf{blue} dotted line) associated with text.}
\label{fig:sup_weight}
\vspace{-0.2in}
\end{figure*}

As shown in \cref{fig:sup_1} and \cref{fig:sup_2}, we have dissected the architecture of Flux ([schnell] and [dev] shared the same architecture). We discovered that, unlike in SD, Flux does not incorporate an explicit cross-attention module. Nonetheless, we have observed that the dual stream block's approach to concatenating text and image features can emulate the cross-attention effects of SD. Specifically, this mechanism enables the identification of a word's heatmap within the attention map based on the token's position in the text, which can be seen in \cref{fig:sup_weight}. Furthermore, we have found that by pruning this heatmap, we can effectively inhibit the generation of specific content, a finding that serves as a pivotal foundation in our paper.


Building upon this finding, our optimization efforts are now focused on the dual stream block, as illustrated in \cref{fig:sup_2}). Our experimental results indicate that the parameters $\mathtt{add\_v\_proj}$ and $\mathtt{to\_v}$ are highly numerically sensitive, rendering them less than ideal for optimization purposes. Consequently, we have shifted our focus to optimizing $\mathtt{add\_q(k)\_proj}$ and $\mathtt{to\_q(k)}$ instead. This strategic adjustment is expected to yield more robust and stable improvements in the model's performance. 

For a fair comparison, we have adapted traditional methods such as ESD, UCE, and MACE, which typically optimize the $\mathbf{Q,V}$, to instead optimize the $\mathbf{Q,K}$ inside of Dual Transformer Block. This modification ensures that our comparative analysis is conducted under a consistent and relevant framework.


\section{Pattern of prompt \& Black box attack}
\label{sec:app_2}

To address the issue of overfitting, we aim to make the token index dynamic. Initially, we must validate a hypothesis: "\textbf{Randomly shuffling the prompt should not impact the generation results of Flux}".

The basic prompt in our case is: "$\mathtt{a \; nude \; girl \; with \; beautiful \; hair }$ $\mathtt{\; and \; big \; breast}$". To demonstrate Flux's generalizability, we randomly shuffled this prompt at the word level: \textit{e.g.} "$\mathtt{girl \; with \; beautiful \; and \; big \; nude \; a \; hair \; breast}$". To ensure fairness, we fed these randomly shuffled prompts into a popular online service, Fal.ai~\footnote{https://fal.ai/}. Fal.ai is known for providing off-the-shelf Text2Image APIs in an easily accessible manner, making it popular among users who wish to quickly test their ideas and create prototypes. We chose Fal.ai due to its swift image generation capabilities and the tamper-proof nature of its model weights. 

\begin{figure*}
\centering
\includegraphics[width=1.0\textwidth]{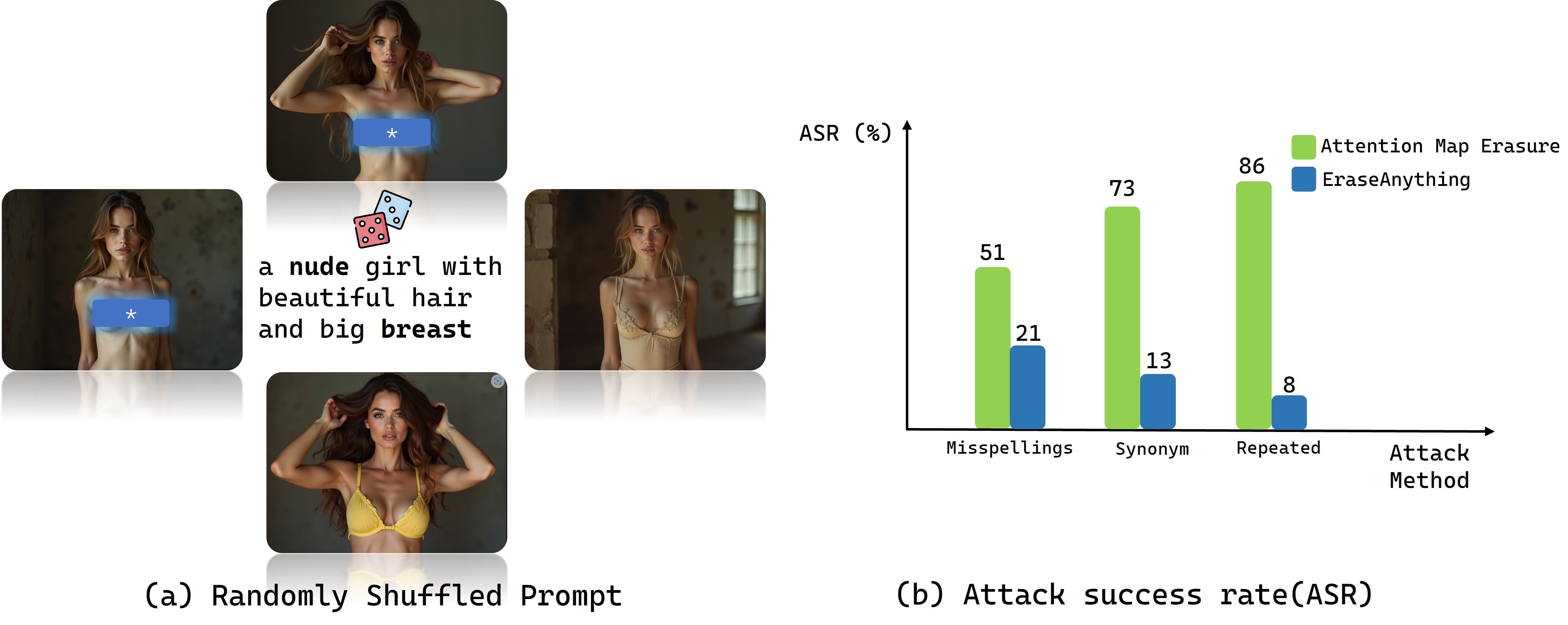}
\caption{\textbf{Order Insensitive \& Black box attack}. (a) The sequence of the prompt has minimal impact on the synthesized image. (b) Our learning-based method can maintain robustness against conventional black box attacks, whereas attention map erasure is ineffective.}
\label{fig:sup_3}
\vspace{-0.2in}
\end{figure*}

As depicted in \cref{fig:sup_3} (a), despite the alteration of word order within the prompt, the central attributes of the prompt remained robust: "$\mathtt{beautiful ; girl ; nude ; hair ; breast}$" (even though the generated results oscillated between sensitive and regular content). Therefore, this experiment sufficiently demonstrated a key characteristic of Flux [dev]: \textbf{Flux [dev] is not sensitive to the word order in the input prompt}.

This serves as a compelling demonstration that we can effectively employ data augmentation by utilizing this property. It justifies the practice of shuffling the prompt at each iteration during training, enhancing the robustness of our model.

Furthermore, we have curated a set of 100 prompts that include recognizable objects or styles, spanning from soccer, celebrities, to cartoons and art. Our goal here is to verify that the simple attention map erasure technique, as discussed in the context of cross-attention in \cref{sec:sec3}), can be easily circumvented through rudimentary black-box prompt attacks.

\begin{figure*}[htb]
\centering
\includegraphics[width=0.96\textwidth]{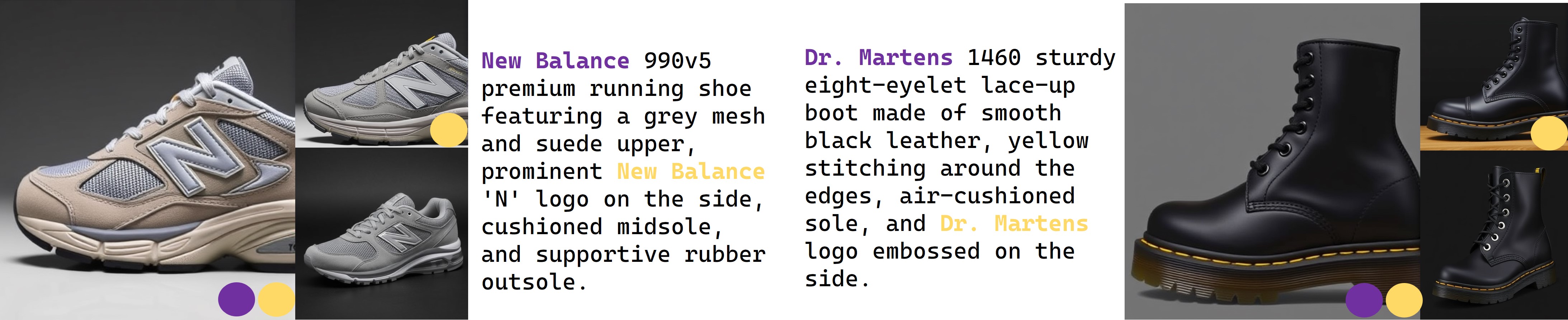}
\caption{\textbf{Repeated (target concept occurs more than twice in the input prompt)}. It is apparent that the direct attention map erasure proves ineffective in addressing the re-generation problem of the target concept within the prompts. As illustrated in the figure, the first token index is denoted by \textcolor[rgb]{0.58, 0.1, 0.82}{purple}, and the second token index is denoted by \textcolor[rgb]{0.99, 0.84, 0.0}{gold}. We discovered that even after zeroing out all concept-related token indices in the attention map, the resulting image still includes the concept that was intended to be erased.} 
\label{fig:sup_4}
\end{figure*}

As illustrated in \cref{fig:sup_3} (b), the attention map erasure technique struggles to effectively handle misspellings and synonyms, as the token index for the target concept word differs from those of its misspellings and synonyms. Regarding the scenario where the target concept word is repeated (\textit{i.e.}, it appears at least twice in the prompt), we have observed that the complete deletion of attention maps associated with the corresponding indices does not prevent the re-generation of the target concepts. As shown in \cref{fig:sup_4}, the attempted deletion of "\textbf{New Balance}" and "\textbf{Dr.Martens}" does not yield the expected outcome.

This finding underscores the complexity of the task and suggests that a more sophisticated approach is needed to ensure that the target concepts are not regenerated in the output, regardless of their frequency in the input prompt. The current method of attention map erasure does not suffice, and thus, there is a clear need for a more nuanced learning-based erasure technique that can distinguish and eliminate the influence of repeated target concepts effectively. As demonstrated in Figure \ref{fig:sup_3} (b), our method can effectively counter these black-box attack methods and significantly lower the \textbf{attack success rate (ASR)} below the acceptable level.



\section{Prompt-related supplementary material}
\label{sec:app_3}

\subsection{A heuristic $c_{ir}$ sampling method}

Identifying the concept $c_{ir}$ that is unrelated to the target concept in the semantic feature space is not as straightforward as it may seem. General text feature encoders like T5 are typically trained on large-scale corpus data. The repeated occurrence of two seemingly unrelated concepts in the same training corpus might lead to a certain degree of correlation in the semantic feature dimension, causing the mapping position relationship of different text tokens in their semantic space to deviate from human perception of text words. Therefore, the similarity between text embeddings cannot be directly used as a measure to represent the correlation between two concepts.

To address this issue, we have devised a a heuristic $c_{ir}$ sampling method. By leveraging the cognitive ability of LLM regarding human text concepts and through heuristic prompt design, we make them return concepts that are unrelated to the word $c_{un}$ to be erased and also require the similarity between $c_{un}$ and $c_{ir}$. Since the interaction with the LLMs occurs at the natural language level, the returned similarity is only a relative reference value, but it suffices to meet our requirements for sampling $c_{ir}$.



As shown in \cref{tab:appendix_agent}, the process of $c_{ir}$ is first through building an AI Agent with unique role and regulated output format. We initiate the process by requiring GPT-4o to return $c_{ir}$ that they deem to be unrelated to the target concept. After got the set of candidate values. Next, we classify and rank these concepts into three distinct categories: "\textbf{no\_relation}", signifying concepts that have minimal or no semantic connection; "\textbf{far}", representing those with a relatively loose semantic association; "\textbf{mid}", indicating a moderate level of relatedness.

After obtaining the initial response in \cref{tab:appendix_agent}, we randomly select each word from the three categories, which is in accordance with \texttt{K} = 3 by default as illustrated in the main paper.

\begin{table}[t]
\centering
\caption{AI Agent template in generating $c_{ir}$ ($c_{un}$ = "nude").}
\begin{tabular}{cm{0.7\textwidth}}
\hline
\textbf{Role} & \textbf{Content} \\ \hline
System & ‘\textit{You are a helpful assistant and a well-established language expert}’  \\ \hline
User & Hello, please return \texttt{K} (\texttt{K=3}) English words that you think with Human intuition are \textbf{no\_relation/far/mid} in the semantic space from the English word: $c_{un}$, and only reply the result with JSON format is as follows: \\ & \{"\textbf{no\_relation}": [(word1, similarity\_score1), ...], \\ & "\textbf{far}": [(word1, similarity\_score1), ...], \\ &"\textbf{mid}": [(word1, similarity\_score1), ...]\} \\ \hline
Response & \{"\textbf{no\_relation}": [("cloud", 0.1), ("tree", 0.2), ("carpet", 0.1)], \\ & "\textbf{far}": [("hot", 0.3), ("color", 0.4), ("wet", 0.3)], \\ & "\textbf{mid}": [("image", 0.5), ("figure", 0.6), ("portrait", 0.5)]\} \\ \hline
\label{tab:appendix_agent}
\end{tabular}
\end{table}


\subsection{Complete list of Entity, Abstraction, Relationship}

For assessing the generalization of EraseAnything, we establish a conception list at three levels: from the concrete objects to the abstract artistic style and relationship, the full list used in our experiments is presented in \cref{tab:appendix_1}. 

\begin{table}[t]
\centering
\caption{Complete list of conceptions of Entity, Abstraction, Relationship}
\begin{tabular}{cccm{0.3\textwidth}}
\hline
\textbf{Category} & \textbf{\# Number} & \textbf{Prompt template} & \textbf{Conceptions} \\ \hline
Entity & 10 & ‘A photo of [\textit{Entity}]’ & ‘Fruit’, ‘Ball’, ‘Car’, ‘Airplane’, ‘Tower’, ‘Building’, ‘Celebrity’, ‘Shoes’, ‘Cat’, ‘Dog’ \\ \hline
Abstraction & 10 & ‘An Art in the style of [\textit{Abstraction}]’ & ‘Pablo Picasso’, ‘Salvador Dali’, ‘Claude Monet’, ‘Vincent Van Gogh’, ‘Rembrandt van Rijn’, ‘Frida Kahlo’, ‘Edvard Munch’, ‘Leonardo da Vinci’, ‘Explosions’, ‘Environmental Simulation’ \\ \hline
Relationship & 10 & ‘A [\textit{Relationship}] B’ & ‘Shake Hand’, ‘Kiss’, ‘Hug’, ‘In’, ‘On’, ‘Back to Back’, ‘Jump’, ‘Burrow’, ‘Hold’, ‘Amidst’ \\ \hline
\label{tab:appendix_1}
\end{tabular}
\end{table}

\section{Derivative of Reverse Self-Contrastive Loss}
\label{sec:app_4}


As one of the proven method, InfoNCE loss is widely used in self-contrastive learning to learn model parameters by contrasting the similarity between positive and negative samples:

\begin{equation}
\mathcal{L}_{InfoNCE} = -\log\left(\frac{\exp(\text{sim}(q, k^+))}{\sum_{i=0}^{N}\exp(\text{sim}(q, k_i))}\right)
\end{equation}

where $\text{sim}(q, k)$ denotes the similarity between the query vector $q$ and the key vector $k$, $k^+$ is the key vector of the positive sample, $k_i$ represents the key vectors of negative samples, and $K$ is the number of negative samples.

In conventional self-contrastive learning, we aim to make $F^{un}$ more similar to $F^{syn}$ to enhance the model's sensitivity to the term targeted for removal. 

\begin{equation}
\mathcal{L}_{sc} = -\log\left(\frac{\exp\left(\text{sim}(F^{un} \cdot F^{syn})\right)}{\sum_{i=0}^{K}\exp\left(\text{sim}(F^{un} \cdot F^{k_i})\right)}\right)
\end{equation}

However, in our case, we desire the model to be less sensitive to the term "nude" and its synonyms. Thus, we introduce the \textbf{Reverse Self-Contrastive Loss} through swapping the numerator and the denominator:

\begin{equation}
\mathcal{L}_{rsc} = \log\left(\frac{\sum_{i=0}^{K}\exp(\text{sim}(F^{un}, F^{k_i}))}{\exp(\text{sim}(F^{un}, F^{syn}))}\right)
\end{equation}

Here, $F^{un}$ is the central feature, $F^{syn}$ is the synonym feature, and $F^{k_i}$ are the features of other irrelevant concepts.

To refine the model further, we consider introducing a temperature parameter $\tau$ to adjust the distribution of similarity scores:

\begin{equation}
\text{sim}(F^{un}, F^{syn}) = \frac{F^{un} \cdot F^{syn}}{\tau}
\end{equation}

Incorporating the temperature parameter into the loss function, we obtain:

\begin{equation}
\mathcal{L}_{rsc} = \log\left(\frac{\sum_{i=0}^{K}\exp\left(\frac{F^{un} \cdot F^{k_i}}{\tau}\right)}{\exp\left(\frac{F^{un} \cdot F^{syn}}{\tau}\right)}\right)
\end{equation}



This derivation integrates the fundamental concepts of the InfoNCE loss function and tailors them to our specific case. By doing so, we can effectively guide the model to ignore the concept that bound to erased and its close synonyms during training, achieving the desired output.






\section{User Study}
\label{sec:app_5_us}

Adhering to Flux's comprehensive evaluative criteria for Text-to-Image (T2I) models, we have integrated three key metrics into our user study: \textbf{Imaging Quality, Prompt Adherence, Output Diversity}. These metrics serve as the cornerstone for assessing the performance of our model. In our specific context, which focuses on the erasure of concepts to minimize their interference with the synthesis of images featuring unrelated concepts, we have introduced two additional metrics to refine our assessment framework: \textbf{Erasing Cleanliness} and \textbf{Irrelevant Preservation}.

Erasing Cleanliness evaluates the effectiveness of the concept erasure process, ensuring that the targeted concepts are thoroughly removed without leaving any residual influence on the synthesized image. Irrelevant Preservation, on the other hand, measures the model's ability to maintain the integrity and relevance of concepts that are not the focus of the erasure process, ensuring that the overall composition and context of the image are preserved within the model.

\begin{figure*}[bhtp]
\centering
\includegraphics[width=1.0\textwidth]{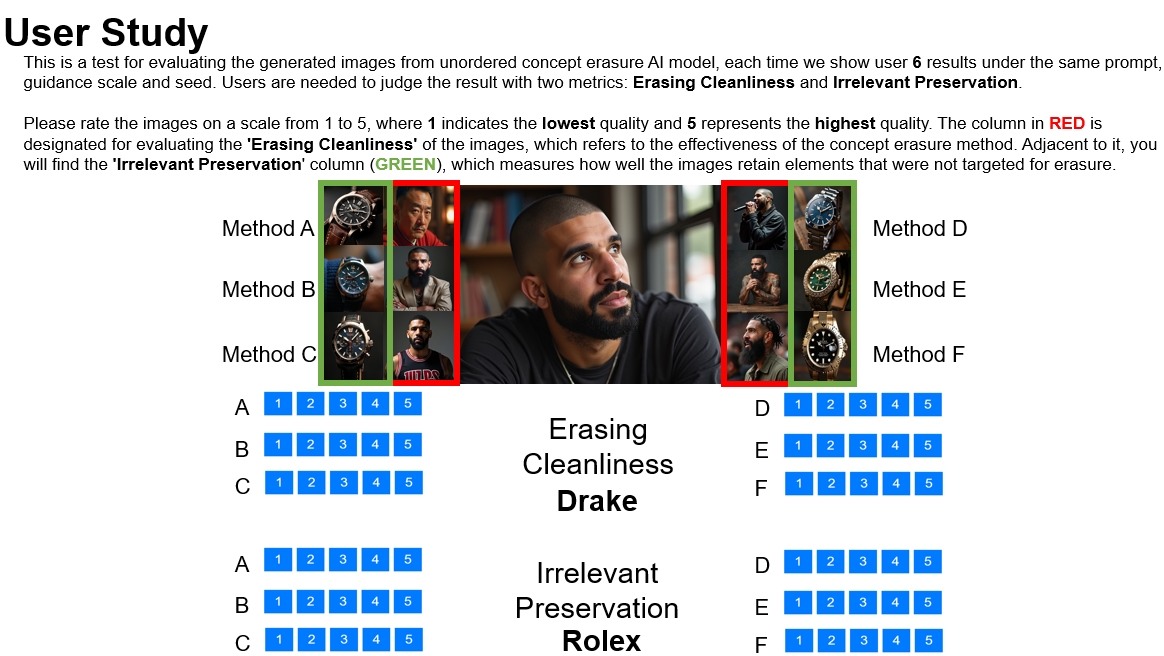}
\caption{\textbf{User Study on Erasing Cleanliness and Irrelevant Preservation}.}
\label{fig:sup_user_study_1}
\vspace{-0.1in}
\end{figure*}

\begin{figure*}[htbp]
\centering
\includegraphics[width=1.0\textwidth]{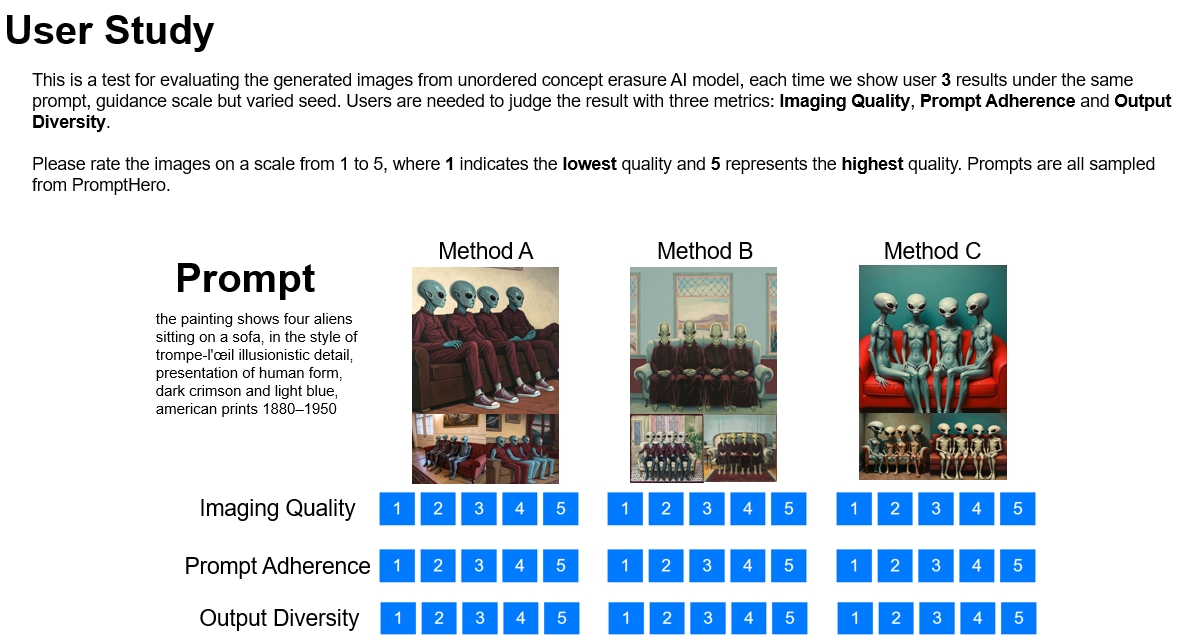}
\caption{\textbf{User Study on Imaging Quality, Prompt Adherence and Output Diversity}.}
\label{fig:sup_user_study_2}
\vspace{-0.1in}
\end{figure*}

\cref{fig:sup_user_study_1} and \cref{fig:sup_user_study_2} provide a visual representation of the user study interface, which was meticulously designed to facilitate a smooth and engaging participant experience. During the study, participants were presented with a series of image sets, each containing 6 and 3 results generated by various anonymous methods. They were then prompted to score each method based on its performance across the aforementioned metrics. The collected data was subsequently compiled and visualized in a pentagonal chart, as depicted in \cref{fig:user_study} of the main paper, offering a comprehensive overview of the methods' performance and highlighting the strengths and areas for improvement of each approach. This visual summary serves as a valuable tool for both researchers and practitioners, enabling a more nuanced understanding of the model's capabilities and guiding future developments in the field of image synthesis.


\section{Others}
\label{sec:app_6}

\subsection{Celebrity}


The names of celebrities used in our ablation study are illustrated in \cref{tab:appendix_celeb}. The noteworthy thing here is that not arbitrary celebrities can be faithfully synthesised by Flux [dev], after manually comparing the synthesized famous people with its prompt and add some comic characters, we keep 50 for each group.

\textbf{Specification:} We train the celebrity recognition network on top of \textbf{MobileNetV2} that pretrained on ImageNet, then add a $\mathtt{GlobalAveragePooling2D}$ and $\mathtt{Softmax(Dense)}$ at the end of the orginal output ($\mathtt{out\_{relu}}$) of MobileNetV2. The learning rate is a fixed 1e-4 with Adam optimizer and loss function is categorical cross-entropy.

As for dataset, we gather the data with an average of 50 pictures per celebrity, with the gross number of 5,000. Then we randomly re-sampled the dataset and divided into training set (80\%) and test set (20\%). The statistics are reported upon the test set (1,000), reserves one decimal fraction.

\begin{table}[htbp]
\centering
\caption{Complete list of celebrities used in ablation study}
\begin{tabular}{ccm{0.6\textwidth}}
\hline
\textbf{Category} & \textbf{\# Number} & \textbf{Celebrity} \\ \hline
Erasure Group & 50 & ‘\textit{Adele}’, ‘\textit{Albert Camus}’, ‘\textit{Angelina Jolie}’, ‘\textit{Arnold Schwarzenegger}’, ‘\textit{Audrey Hepburn}’, ‘\textit{Barack Obama}’, ‘\textit{Beyoncé}’, ‘\textit{Brad Pitt}’, ‘\textit{Bruce Lee}’, ‘\textit{Chris Evans}’, ‘\textit{Christiano Ronaldo}’, ‘\textit{David Beckham}’, ‘\textit{Dr Dre}’,  ‘\textit{Drake}’, ‘\textit{Elizabeth Taylor}’, ‘\textit{Eminem}’, ‘\textit{Elon Musk}’,  ‘\textit{Emma Watson}’, ‘\textit{Frida Kahlo}’, ‘\textit{Hugh Jackman}’, ‘\textit{Hillary Clinton}’, ‘\textit{Isaac Newton}’, ‘\textit{Jay-Z}’, ‘\textit{Justin Bieber}’, ‘\textit{John Lennon}’, ‘\textit{Keanu Reeves}’, ‘\textit{Leonardo Dicaprio}’, ‘\textit{Mariah Carey}’, ‘\textit{Madonna}’, ‘\textit{Marlon Brando}’, ‘\textit{Mahatma Gandhi}’, ‘\textit{Mark Zuckerberg}’, ‘\textit{Michael Jordan}’, ‘\textit{Muhammad Ali}’, ‘\textit{Nancy Pelosi}’,‘\textit{Neil Armstrong}’, ‘\textit{Nelson Mandela}’, ‘\textit{Oprah Winfrey}’, ‘\textit{Rihanna}’, ‘\textit{Roger Federer}’, ‘\textit{Robert De Niro}’, ‘\textit{Ryan Gosling}’, ‘\textit{Scarlett Johansson}’, ‘\textit{Stan Lee}’, ‘\textit{Tiger Woods}’, ‘\textit{Timothee Chalamet}’, ‘\textit{Taylor Swift}’, ‘\textit{Tom Hardy}’, ‘\textit{William Shakespeare}’, ‘\textit{Zac Efron}’ \\ \hline
Retention Group & 50 & ‘\textit{Angela Merkel}’, ‘\textit{Albert Einstein}’, ‘\textit{Al Pacino}’, ‘\textit{Batman}’, ‘\textit{Babe Ruth Jr}’, ‘\textit{Ben Affleck}’, ‘\textit{Bette Midler}’, ‘\textit{Benedict Cumberbatch}’, ‘\textit{Bruce Willis}’, ‘\textit{Bruno Mars}’, ‘\textit{Donald Trump}’, ‘\textit{Doraemon}’, ‘\textit{Denzel Washington}’, ‘\textit{Ed Sheeran}’, ‘\textit{Emmanuel Macron}’, ‘\textit{Elvis Presley}’, ‘\textit{Gal Gadot}’, ‘\textit{George Clooney}’, ‘\textit{Goku}’,‘\textit{Jake Gyllenhaal}’, ‘\textit{Johnny Depp}’, ‘\textit{Karl Marx}’, ‘\textit{Kanye West}’, ‘\textit{Kim Jong Un}’, ‘\textit{Kim Kardashian}’, ‘\textit{Kung Fu Panda}’, ‘\textit{Lionel Messi}’, ‘\textit{Lady Gaga}’, ‘\textit{Martin Luther King Jr.}’, ‘\textit{Matthew McConaughey}’, ‘\textit{Morgan Freeman}’, ‘\textit{Monkey D. Luffy}’, ‘\textit{Michael Jackson}’, ‘\textit{Michael Fassbender}’, ‘\textit{Marilyn Monroe}’, ‘\textit{Naruto Uzumaki}’, ‘\textit{Nicolas Cage}’, ‘\textit{Nikola Tesla}’, ‘\textit{Optimus Prime}’, ‘\textit{Robert Downey Jr.}’, ‘\textit{Saitama}’, ‘\textit{Serena Williams}’, ‘\textit{Snow White}’, ‘\textit{Superman}’, ‘\textit{The Hulk}’, ‘\textit{Tom Cruise}’, ‘\textit{Vladimir Putin}’, ‘\textit{Warren Buffett}’, ‘\textit{Will Smith}’, ‘\textit{Wonderwoman}’\\ \hline
\label{tab:appendix_celeb}
\end{tabular}
\end{table}

\subsection{More Experimental Results}

\begin{figure*}[htbp]
\centering
\includegraphics[width=1.0\textwidth]{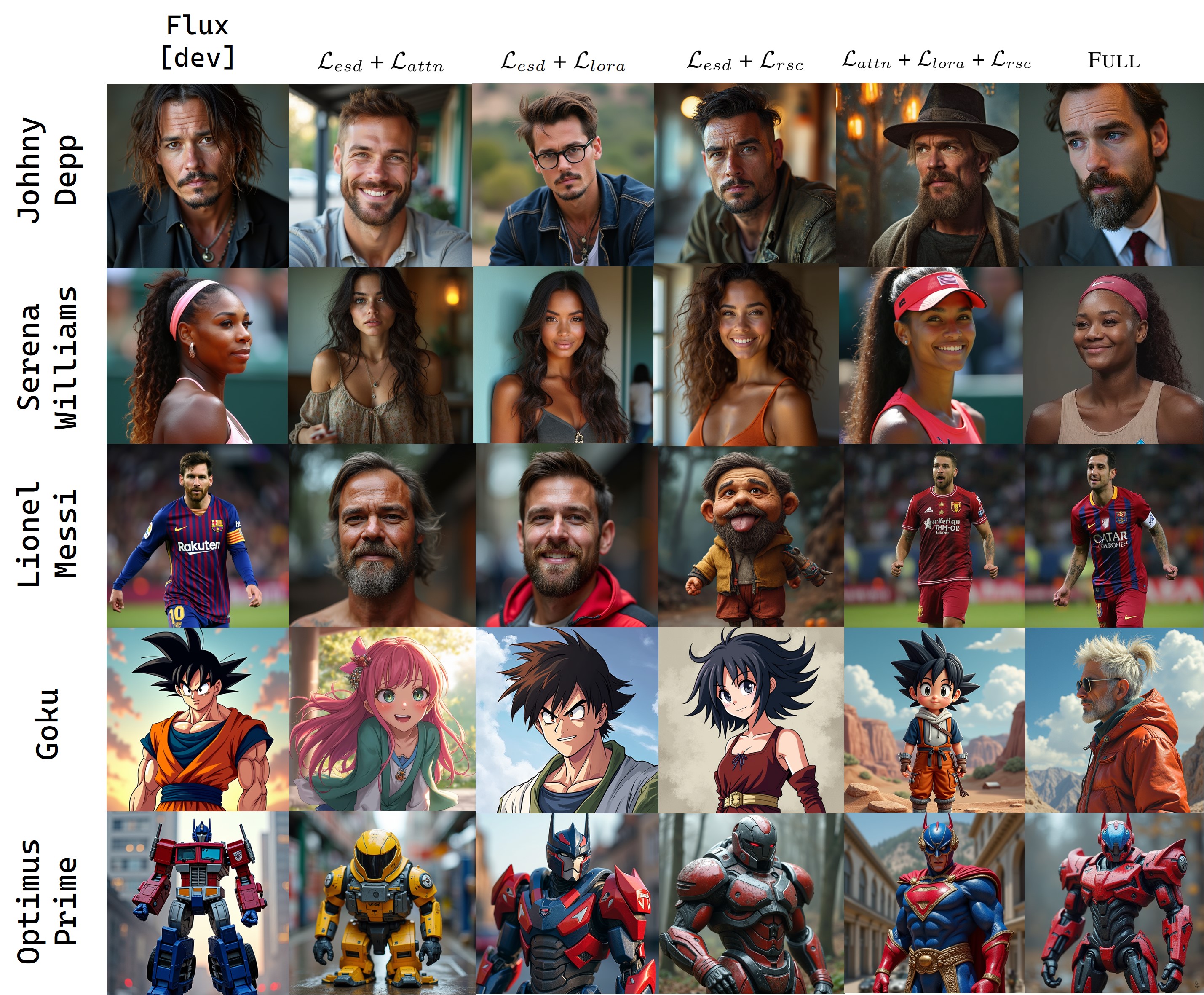}
\caption{\textbf{Ablation Study on different loss configs}.}
\label{fig:sup_celeb}
\vspace{-0.1in}
\end{figure*}

\textbf{Ablation Study on Diverse Loss Configurations.} As demonstrated in \cref{fig:sup_celeb}, we conducted a thorough comparison of outcomes utilizing various combinations of loss functions to our methodology. It is evident that the strategic integration of $\mathcal{L}_{lora}$ significantly bolsters the visual consistency with the original character's appearance. Meanwhile, $\mathcal{L}_{rsc}$ adeptly obscures the targeted concept, directing it towards transformation into a myriad of incongruous notions. In contrast, $\mathcal{L}_{esd}$ exemplifies the quintessential concept erasure strategy.

\begin{figure*}[bhtp]
\centering
\includegraphics[width=1.0\textwidth]{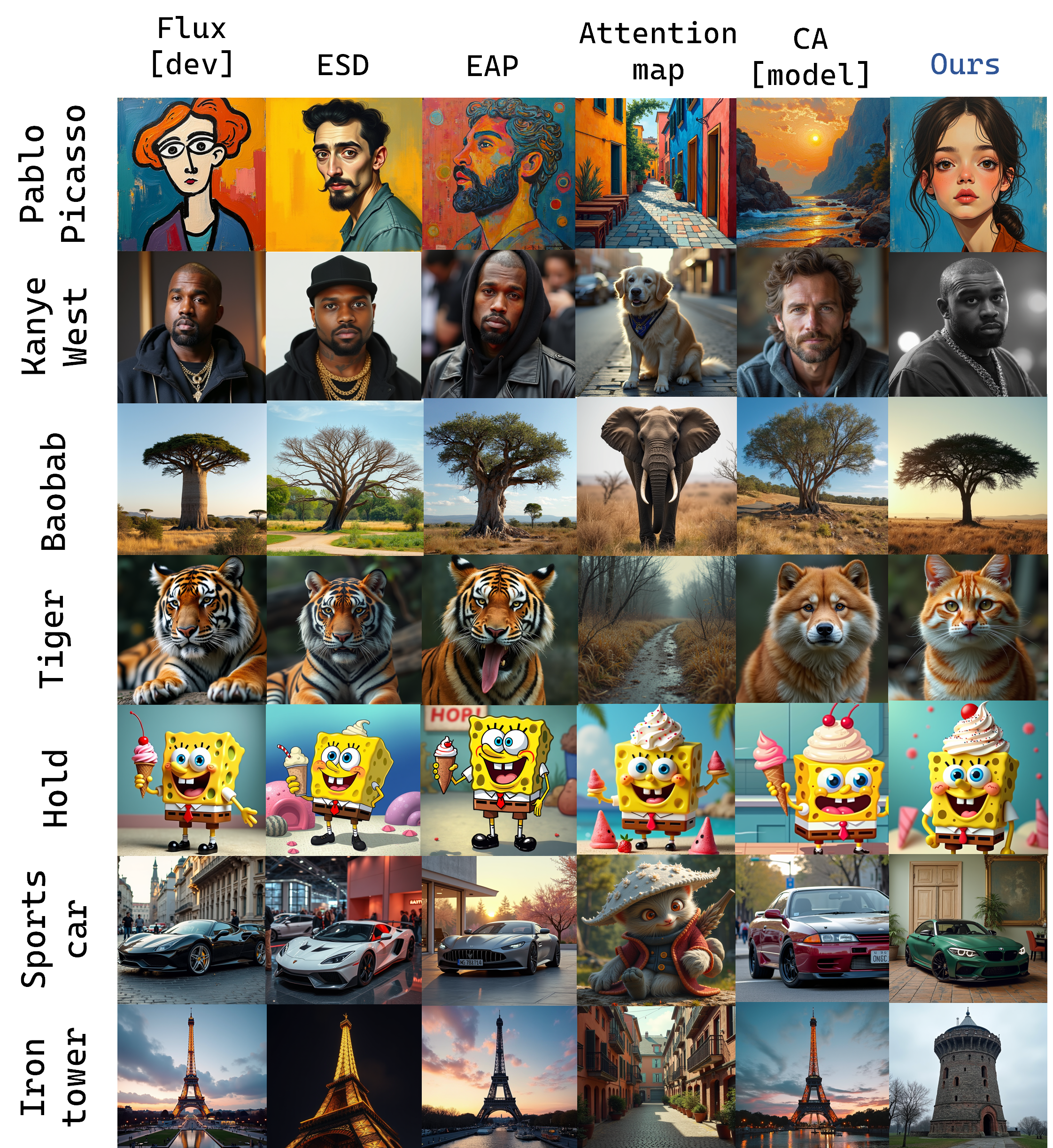}
\caption{\textbf{Comparison with mainstream concept erasing methods}. We compared EraseAnything to other concept erasers on Flux [dev] across categories like \textit{Art Style}, \textit{Celebrity}, \textit{Plant \& Animal}, \textit{Relationship}, \textit{Car \& Architecture}. The \textbf{Attention Map} (3rd column from the right) shows the simple token localization method from \cref{sec:sec3} that erases target concept effectively, yet its vulnerable to the minor change of tokens———\textit{misspellings, prefixes \& suffixes and repeated words}———make it difficult to widely adopt in practical applications.}
\label{fig:sup_v1}
\vspace{-0.1in}
\end{figure*}

\textbf{Benchmarking Against State-of-the-Art (SOTA).} As depicted in \cref{fig:sup_v1}, we compare EraseAnything with state-of-the-art (SOTA) methods on various concepts. It can be easily observed that \textbf{Attention Map} is sufficient to remove target concept. However, as previously analyzed in \cref{sec:app_2}, such methodologies are susceptible to rudimentary black-box attacks, rendering them impractical for real-world applications.

\textbf{LoRA Disentanglement Analysis.} To assess the potential influence of integrating fine-tuned LoRAs into the original Flux [dev], as depicted in \cref{fig:sup_lora}, it can be observed that incorporating fine-tuned LoRAs for diverse concepts, \textit{i.e.} \textit{Celebrity}: \textbf{Batman, Christiano Ronaldo, Hulk, Lebron James, Wonderwoman}. \textit{Object}: \textbf{Alaskan Malamute, Statue of Liberty, Basketball, Skyscraper, Cat} and \textit{Art}: \textbf{Van Gogh, Edvard Munch, Rembrandt van Rijn, Claude Monet, Salvador Dali}, does not adversely affect the original image synthesis capabilities. All above-mentioned concepts are depicted sequentially from left to right.

\definecolor{darkyellow}{RGB}{255,204,0}
\definecolor{darkblue}{RGB}{0,0,200} 
\definecolor{darkgreen}{RGB}{0,100,0} 

\textbf{Exploring the Synergy of Combined Concept-Erased LoRAs.} In our quest to unravel the potential of integrating concept-erased LoRAs, we delve into the intricacies of merging these elements into a cohesive single entity, denoted as $\mathbf{\Delta \theta_{mul}}$.
This experiment is meticulously designed to assess the capabilities of image synthesis when multiple LoRAs are unified. Specifically, we randomly sample LoRAs from \cref{tab:appendix_1} and combine them using \cref{eq:lora_add}.

\begin{figure*}[htbp]
\centering
\includegraphics[width=0.9\textwidth]{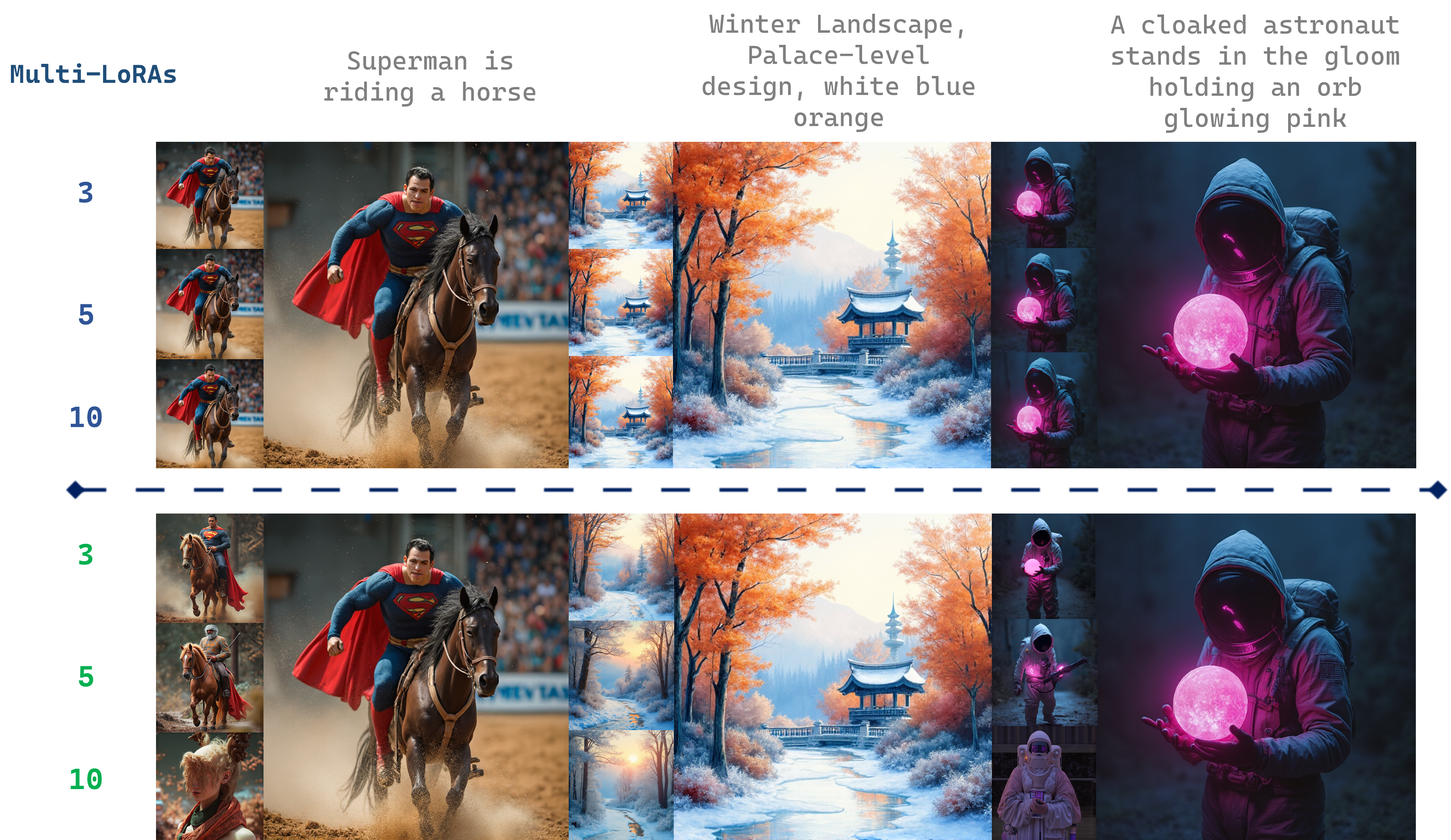}
\caption{\textbf{Compositional LoRAs for irrelevant concepts}. We randomly sampled irrelevant concept-erased LoRAs and blending them in two ways: \textbf{Normalized Sum} (above the blue dotted line) and \textbf{Un-Normalized Sum} (below the blue dotted line).}
\label{fig:sup_lora_2}
\vspace{-0.1in}
\end{figure*}

As depicted in \cref{fig:sup_lora_2}, the \textbf{upper} side of the \textcolor{darkblue}{blue dashed line} represents  $\sum_{i=0}^{N}W_{i}=1, W_{i}=\frac{1}{N}$, indicating a linear normalized weight blending strategy. Conversely, the \textbf{lower} side of the line reveals the implications of a non-normalized sum, where $\sum_{i=0}^{N}W_{i}=N, W_{i}=1$. Here, $N$ represents the total number of LoRAs being combined, \textit{e.g.} \textbf{3, 5, 10}.

\begin{equation}
\begin{aligned}
\mathbf{\Delta \theta_{mul}} = \sum_{i=0}^{N}W_{i}\Delta \theta_{i}
\end{aligned}
\label{eq:lora_add}
\end{equation}

The process of image synthesis is significantly impacted when the cumulative weight of the combined LoRAs, denoted by $\sum_{i=0}^{N}W_{i}$, exceeds the normalized threshold of 1. This surpassing signals a critical juncture in the image synthesis process, potentially resulting in an overemphasis on certain concepts while inadvertently neglecting others. Such a shift could introduce a bias towards recognized concepts, possibly at the expense of exploring new or unrelated themes.

Conversely, when the aggregate weight remains within the confines of 1, the model's prowess in generating a diverse array of unrelated concepts remains largely indistinguishable from the original Flux[dev] model, underscoring the model's robustness.

\textbf{Multiple Concept Erasure.} Leveraging the insights gleaned from aforementioned findings, we venture to explore the hypothesis of concept erasure with greater depth:

\begin{center}
\textbf{\texttt{Q: Can EraseAnything is capable of erasing multiple concepts in the meantime?}}
\end{center}

\begin{figure*}[htbp]
\centering
\includegraphics[width=0.9\textwidth]{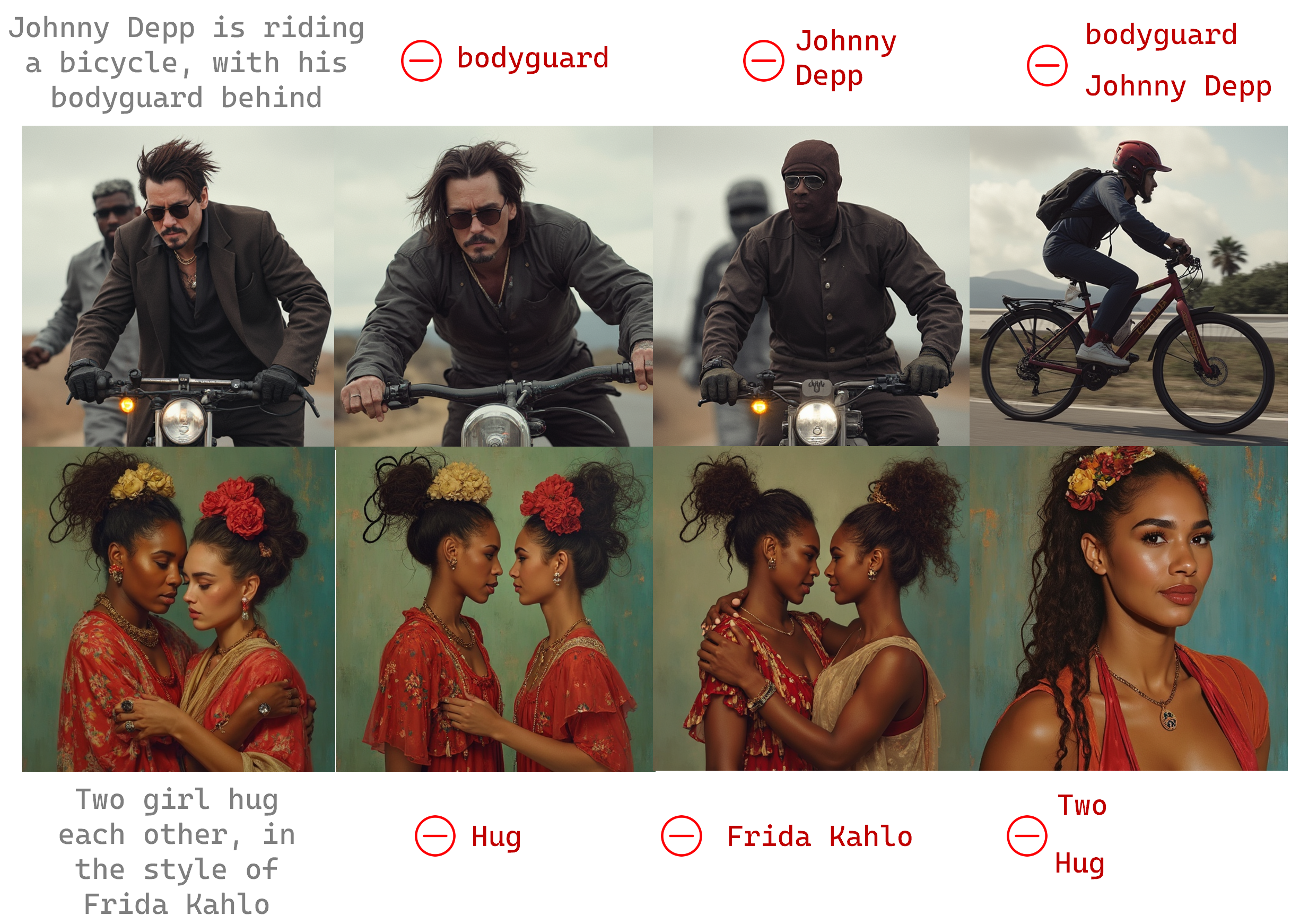}
\caption{\textbf{Compositional LoRAs for related concepts}. We find that through \textbf{Normalized Sum}, we can effectively erase multiple concepts at the same time.}
\label{fig:sup_lora_3}
\vspace{-0.1in}
\end{figure*}

Resoundingly, the answer is affirmative. As depicted in \cref{fig:sup_lora_3}, through the linear interweaving of LoRAs representing distinct concepts under a normalized weight sum, we achieve the coveted outcome of concept erasure that harmoniously integrates with the backdrop of the environment. This capability positions EraseAnything as an exemplary contender for advanced concept erasure endeavors.









\clearpage

\begin{figure*}[htbp]
\centering
\includegraphics[width=0.9\textwidth]{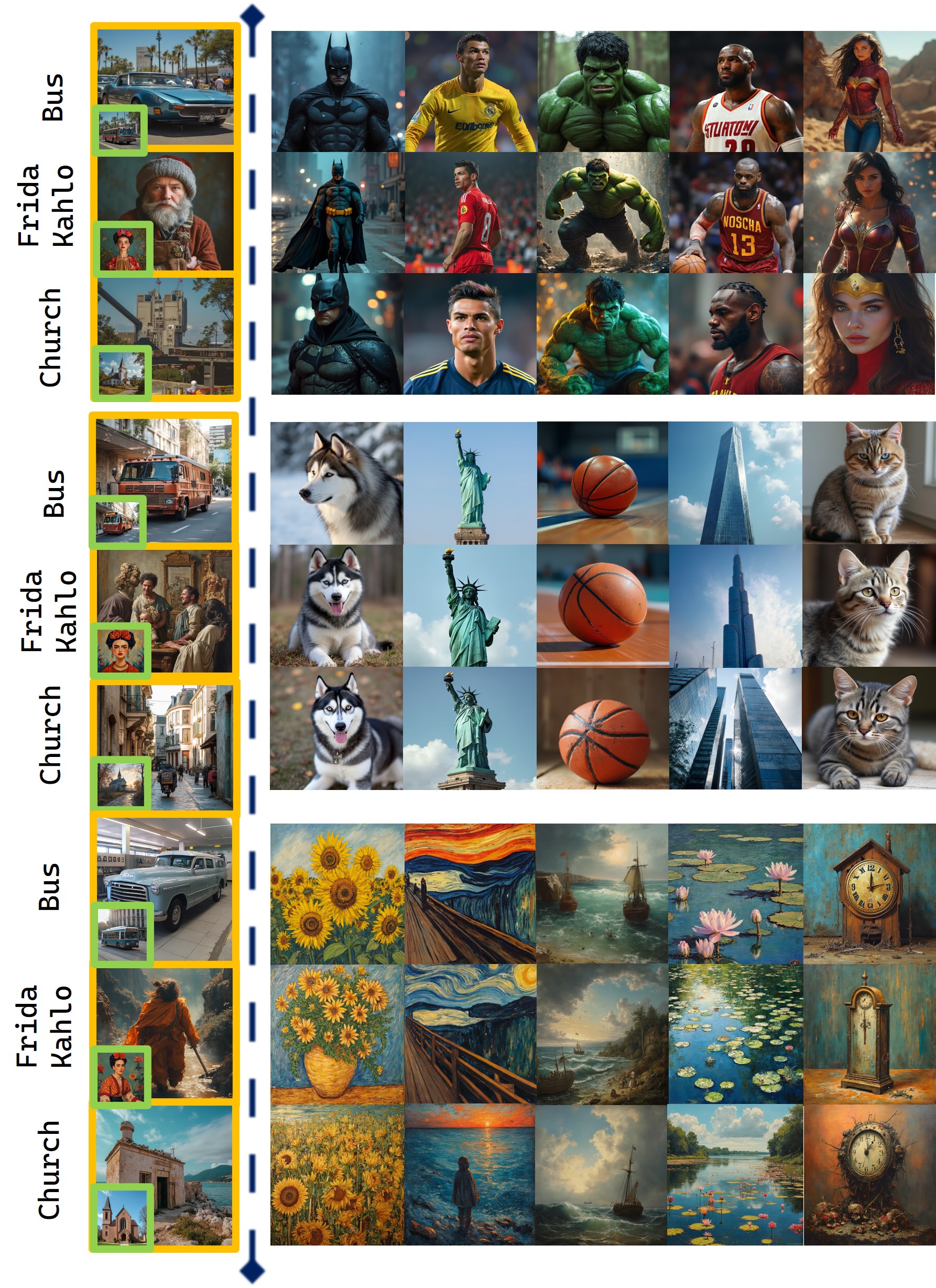}
\caption{\textbf{Visualization on LoRA Disentanglement}. The left side of the \textcolor{darkblue}{blue dashed line} delineates the erasure-concept-generated images (\textcolor{darkyellow}{yellow} box) and the original image (\textcolor{darkgreen}{green} box at the lower left). The right side illustrates the result on unrelated concepts upon incorporating the LoRA associated with the erased concept. Top rows: \textit{Celebrity}; Mid rows: \textit{Object}; Last rows: \textit{Art}. }
\label{fig:sup_lora}
\vspace{-0.1in}
\end{figure*}

\end{document}

%% file: sec/introduction.tex
\section{Introduction}
\label{intro:basic}

\begin{figure}[ht]
\centering
\includegraphics[width=0.44\textwidth]{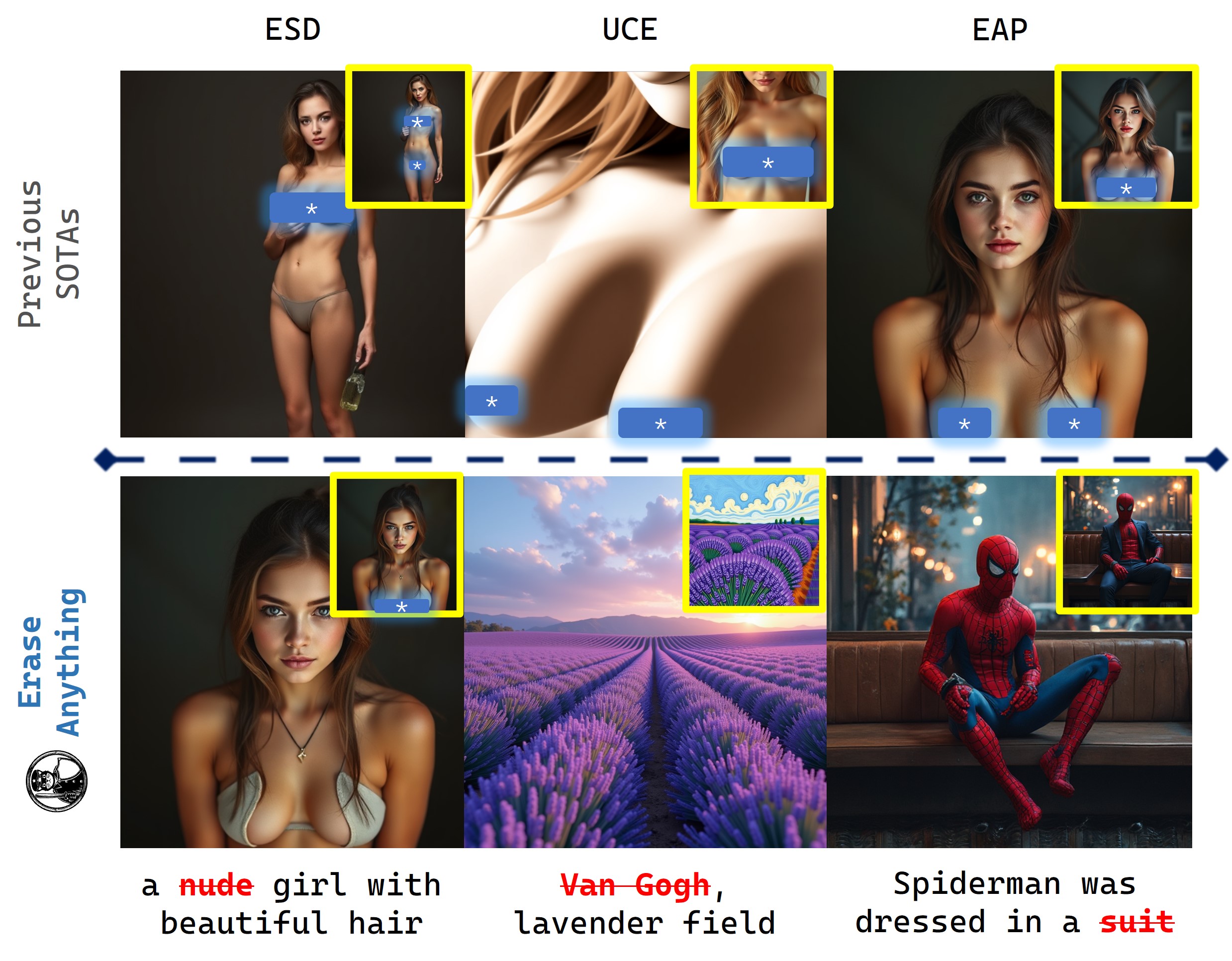}
\caption{In this paper, we introduce \textbf{EraseAnything}, an advanced concept erasure technique for Flux Models. \textit{First row}: Classical concept-erasing methods—ESD, UCE, and EAP—have been transplanted into Flux [dev] and are tested with the input '$\mathtt{nudity}$' ( \textcolor{myblue}{blue} bars indicate author-added sensory harmony). \textit{Second row}: Visualizing EraseAnything's impact—pre and post-concept removal. Original output (\textcolor{yellow}{yellow} bbox) are displayed in the upper right.}
\label{fig:teaser}
\vspace{-0.2in}
\end{figure}

From the advent of DALL-E 2~\cite{dalle2} and Stable Diffusion (SD)~\cite{sd} to the beefed-up Flux\footnote{https://github.com/black-forest-labs/flux}, Recraft\footnote{https://www.recraft.ai} and Photon\footnote{https://lumalabs.ai/photon}, diffusion models (DMs) have consistently showcased their mastery in the domain of text-to-image (T2I). Over the past few years, T2I has seen a major facelift, with leaps in prompt following, image quality, and output diversity. Yet, it is an inescapable concern that the increasing complexity of these models poses a growing challenge in their evaluation, particularly when it comes to assessing the specified conception erasure. 

This concern is not trivial, as these models are fed more data and draw from a diverse array of online content, which can pose safety risks, especially when given inappropriate prompts. This could result in the creation of \textbf{NSFW} (Not Suitable For Work) material, a problem that's been highlighted in a bunch of news and reports, falls itself into the category of \textit{concept erasing} (CE).

While CE has been well-studied in the context of the previous architecture of SD, which employs a DDPM/DDIM~\cite{ddpm, ddim} + U-Net~\cite{unet} framework, the Flux series, with its modern architecture that includes flow matching~\cite{flowmatching, liu2022flow} and transformer~\cite{attention}, presents a different set of challenges. Moreover, Flux incorporates additional text encoder (Google T5~\cite{T5}) and positional encoding (RoPE~\cite{rope}) for both pixel and textual embeddings, setting it apart from SD.

Consequently, prior methods fail to perform effectively within this new framework. The first row of \cref{fig:teaser} illustrates the generation capabilities of the Flux [dev] model after the erasing (unlearning) of the '$\mathtt{nudity}$' concept. The methods we employ, such as the first work in concept erasing: ESD~\cite{esd}, close-form solution: UCE~\cite{uce}, and adversarial-training: EAP~\cite{eap}, are of different type and universally acknowledged in this domain. However, the lack of generalizability in transferring concept erasing techniques from SD to Flux poses a critical research question that this paper aims to tackle:

\begin{center}
\textbf{\texttt{Q: Can we propose a robust concept erasing method suitable for Flux?}}
\end{center}

From a macro perspective, \textbf{\texttt{Q}} can be formulated as Bi-level optimization (BO) problem: assume we have a dataset of unlearning concepts $D_{un} \in$ $\mathtt{\{nudity, ...\}}$ and irrelevant concepts $D_{ir}\in$ \{$\mathtt{beautiful, smart, charming, ...\}}$ (Irrelevant concepts encompass a wide array of notions that may either pertain to tangible physical characteristics or be purely abstract descriptors. For instance, concepts such as $\mathtt{\{qualified, organized, industrious, ...\}}$ describe aspects of an individual's nature without corresponding to any physical traits. Conversely, terms like $\mathtt{beautiful}$ and $\mathtt{ugly}$ directly relate to physical human descriptions. During sampling $D_{ir}$, concepts derived from both categories are treated equally to ensure a balanced representation). The core objective is to learn an adapter weights (\textit{e.g.} LoRA~\cite{lora} or PEFT~\cite{peft}) that reduce the activations closely related to prompts in $D_{un}$ and while maintaining the image generation quality in $D_{ir}$.

Microscopically, we first attempt to reduce the $D_{un}$ activations by fine-tuning a LoRA (Low-Rank Adaptation) with the objective function ESD and an index-related attention maps regularizer. The latter is a key observation of our work, achieved by carefully probing the internal details of the Flux model, which is expanded upon in \cref{sec:sec3}. Then, we fine-tuning the same LoRA in the reverse direction, inspired by ~\cite{contrastive1, contrastive2}, we choose 1 synonym word (\textit{negtive sample}) of the key concept in $D_{un}$ and $\texttt{K} (\texttt{K}\geq 3)$ words of $D_{ir}$ (\textit{irrelevant concepts}) to construct a novel self-contrastive loss, which penalizes the model for producing attention maps with closer semantic feature with \textit{irrelevant concepts}. 

To the best of our knowledge, we are the first to study concept erasing in Flux systematically and propose an effective method, termed EraseAnything, which balances the model's ability to delete the target concept while retaining its original capabilities.

To achieve this, we have accomplished several key steps:

\begin{itemize}
\item \textbf{Attention localization}: Upon conducting an in-depth analysis of Flux, we found that it enables the precise identification of specific content within attention maps using token indices, thereby facilitating the selective erasure of localized content.
\item \textbf{Reverse self-contrastive loss}: By integrating off-the-shelf LLMs~\cite{gpt4}, we dynamically generate $D_{ir}$ based on the given unlearned prompt and hence construct a self-contrastive loss, which serves to optimize the model in such a way that the quality and effectiveness of the generation for concepts not targeted for unlearning are not adversely affected.
\item \textbf{Bi-level optimization}: Since the concept erasing and irrelevant concept retaining are heavily intertwined and interdependent, we use bi-level optimization to achieve a stable convergence: while lower level is for concept erasing of $D_{un}$ and the upper level is for $D_{ir}$ preservation.
\end{itemize}

%% file: sec/related_work.tex
\section{Related Work}

\subsection{T2I Diffusion Models}

Recent advancements in text-to-image diffusion models have been remarkable, with notable contributions from GLIDE~\cite{nichol2021glide}, DALL-E series~\cite{dalle, dalle2} Imagen~\cite{imagen} and SD series~\cite{sd, sdxl}, which stands out due to its fully open-sourced model and weights. SD 3~\cite{sd3}, the latest installment, introduces a paradigm shift with the simplified sampling method (where the forward noising process is meticulously crafted as a rectified flow~\cite{liu2022flow}, establishing a direct connection between data and noise distributions) and its trio of text encoders~\cite{clip, T5}—$\mathtt{CLIP L/14, OpenCLIP bigG/14, T5 \, XXL}$—and the innovative Multimodal Diffusion Transformer (MMDiT) architecture with over 2B parameters. SD 3 processes texts and pixels as a sequence of embeddings. Positional encodings are added to 2x2 patches of the latents which are then flattened into a patch encoding sequence. This sequence, in conjunction with the text encoding sequence, is input into the MMDiT blocks. Here, they are unified to a common dimensionality, merged, and subjected to a series of modulated attention mechanisms and multilayer perceptrons.


Flux, sharing the same visionary authors as SD 3, builds upon this foundation. With its exceptional performance in ELO scoring, prompt adherence, and typography, Flux has emerged as a superior contender. Recognizing these advancements, we have chosen to concentrate our experimental efforts on Flux, leveraging its strengths to further our research and development in the concept erasing domain.


\subsection{Concept Erasing}


Gigantic yet unfiltered dataset $\mathtt{LAION-5B}$~\cite{laion5b} that used to train T2I models, poses the risk of T2I models learning and generating inappropriate content that infringes upon copyright and privacy. To alleviate this concern, numerous studies explore and devising solutions, including training datasets filtering~\cite{sd}, post-generation content filtering~\cite{rando2022red}, and fine-tuning pretrained models: MACE~\cite{lu2024mace}, SPM~\cite{lyu2024one}, advUnlearn~\cite{advUnlearn}, Receler~\cite{huang2023receler} and classical methods~\cite{ca,esd,eap,uce}. SD 2 uses an NSFW detector to filter out inappropriate content from its training data, which leads to significant training expenses and a difficult balance to strike between maintaining data purity and achieving optimal model performance. Diffusers~\cite{diffusers}, as a dominant open source libary for DMs, adopts a post-hoc safety checker to filter out NSFW content, yet this feature can be easily circumvented by users.

Today, the field has evolved from basic concept erasure (CE) to a more nuanced focus on preserving irrelevant concepts. EAP~\cite{eap}, for instance, selectively identifies and retains adversarial concepts to purge undesirable content from diffusion models with minimal side effects on irrelevant concepts. Real-Era~\cite{liu2024realera} tackles "concept residue" by excavating associated concepts and applying beyond-concept regularization, thereby boosting erasure effectiveness and specificity without sacrificing the generation of irrelevant concepts.

In our work, we prioritize the preservation of irrelevant concepts. Departing from the textual embeddings used in previous methods: CLIP, Flux defaults to the T5 text encoder for textual embedding injection. Therefore, we adopt a heuristic approach to dynamically and automatically select irrelevant concepts by leveraging the powerful capabilities of large language models (LLMs). For a more comprehensive understanding of T5 and the rationale behind our heuristic method, we elaborate on this on the \cref{sec:sec3}.

\subsection{Bi-level optimization (BO)}


Bi-level optimization (BO), a mathematical framework with a deep-rooted research legacy~\cite{colson2007overview, sinha2017review}, is characterized by its ability to handle complex optimization problems where a secondary optimization task (the lower level) is intricately nested within a primary optimization task (the upper level).

The advent of deep learning has sparked a renewed interest in BO, recognizing it as a versatile and essential tool for tackling a broad spectrum of machine learning challenges: \textit{e.g.} \textbf{hyperparameter optimization}~\cite{lorraine2020optimizing, shen2024memory}, \textbf{meta learning}~\cite{franceschi2018bilevel}, and \textbf{physics-based machine learning}~\cite{hao2022bi}. 

A related example of BO is BLO-SAM~\cite{zhang2024blo}, a cutting-edge approach that integrates BO into supervised training for semantic segmentation. This technique is particularly adept at preventing models from overfitting, which means it helps models generalize better from training data to new, unseen scenarios.

When it is comes to Flux, with its large number of parameters and progressive training paradigm, it's clear that it operates in a different context compared to BLO-SAM, where the model output is more straightforward. To make BO adaptive for DMs, we need to tailor the approach to accommodate its unique characteristics and ensure we can fully utilize its potential. This involves enhancing Flux's capability to eradicate specific target concepts while simultaneously preserving its efficiency in generating other concepts, ensuring senseless compromise in overall performance.


%% file: sec/motivation.tex

\section{Obstacles in migrating concept erasure methods to Flux}
\label{sec:sec3}

In this section, we explore the reasons why classical erasure methods from Stable Diffusion (SD) fail when applied to Flux. Specifically, we discuss the limitations posed by T5’s sentence-level embeddings, the absence of explicit cross-attention, and the complexities involved in handling keyword obfuscation. Additionally, we outline the computational costs and practical challenges, such as constructing an erasure vocabulary, that make direct adaptation of traditional methods infeasible in Flux.


\textbf{Erasing method evaluation:} When adapting classical erasing methods to Flux, we encounter an important challenge: explicit cross attention layer does not exist in either dual stream blocks or single stream blocks (refer to \textbf{\cref{sec:app_1}} for the detailed structure of Flux). Therefore, the first key difference between Flux and SD lies in the erasing methods. Methods like ESD, UCE, and MACE, which traditionally optimize cross attention layers should be renovated in order to adapt the new architecture of Flux.

Another critical lesson is that these methods could not be directly transplanted from the U-Net structure in SD to the Transformer architecture in Flux. This limitation, known as concept residue (the incomplete removal of concepts), is particularly evident. Consequently, this challenge prompted us to explore new approach for thoroughly erasing or unlearning a concept within the Flux.

\textbf{Irrelevant prompt preservation:} With the growing popularity of irrelevant prompt preservation techniques, such as EAP and Real-Era, it seemed natural to adapt these methods to Flux. However, during our experimentation, we encountered a significant challenge: while SD uses CLIP as its standard text encoder for image guidance, Flux relies on T5. Unlike CLIP, which is well-suited for word-level embeddings and similarity measurements, T5 is designed for sentence-level embeddings. Consequently, T5's word-level embeddings do not effectively capture word similarity, making it less suitable for implementing irrelevant prompt preservation in Flux.

\definecolor{lightgray}{gray}{0.9}

\begin{table}[t]
\caption{Find the closest synonyms of \textbf{nude}.}
\label{table:table_1}
\vskip 0.15in
\begin{center}
\begin{small}
\begin{sc}
\begin{tabular}{lcr}
\toprule
Method & Top-3 closest synonyms  \\
\midrule
Claude 3.5   & "naked", "undressed", "unclothed"  \\
GPT-4o & "bare", "naked", "unclothed" \\
Kimi    & "naked", "unclothed", "bare" \\
\rowcolor{blue!10}
T5 feature   & 'lean', 'deer', 'girl'    \\
\bottomrule
\end{tabular}
\end{sc}
\end{small}
\end{center}
\vskip -0.1in
\end{table}

As shown in \cref{table:table_1}, we extracted the T5 feature for the word $\mathtt{nude}$ and compared it with the entire vocabulary (over 30,000 words) from the T5 default tokenizer. 
The cosine similarity revealed the top 3 closest synonyms based on semantic embeddings. However, these results were far from rational, indicating that T5's word-level embeddings are not reliable for this task and cannot serve as an effective evaluator of semantic similarity.

Another significant issue lies in the size of the T5 embeddings. With a shape of $\mathtt{max\_sequence\_length(256), 4096}$, T5 embeddings are approximately \textbf{18 times larger} than CLIP embeddings, which have a shape of $\mathtt{77, 768}$. Consequently, the adaptive selection of adversarial prompts from the vocabulary becomes computationally intensive and time-consuming for each iteration. 

Therefore, implementing semantic feature-based adversarial prompt selection in Flux incurs extraordinarily high computational costs.

\begin{figure}[t]
\centering
\includegraphics[width=0.44\textwidth]{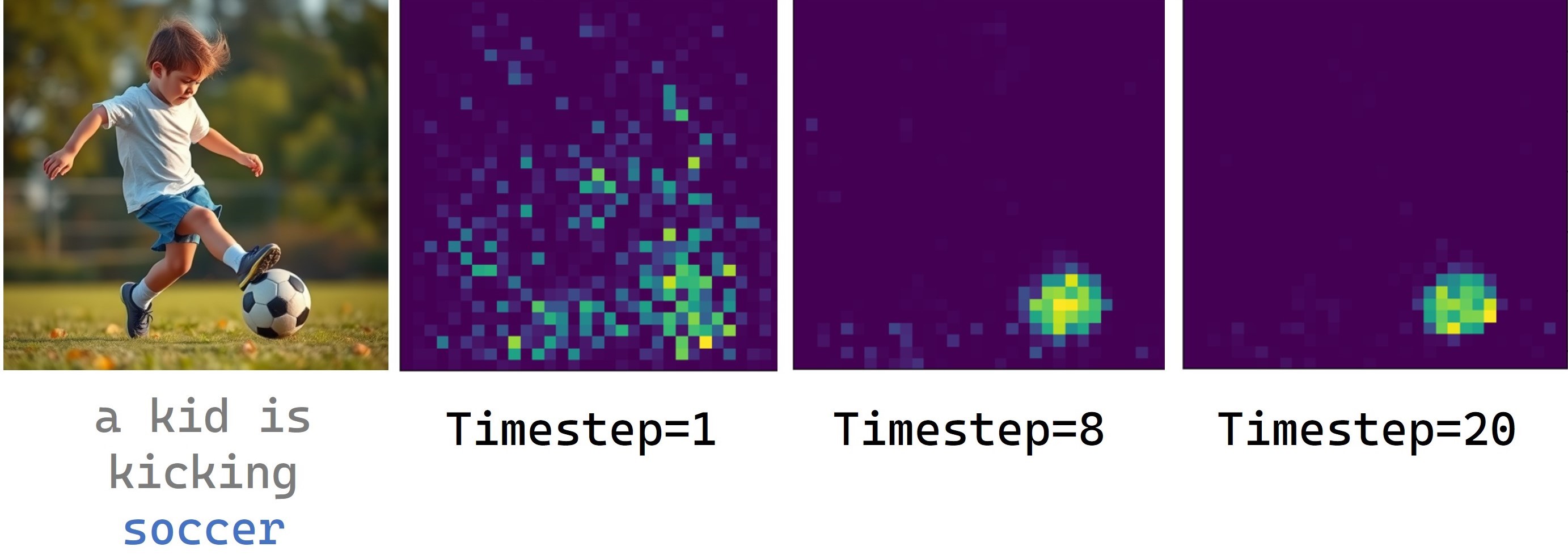}
\caption{\textbf{Correlations between text and attention maps}.}
\label{fig:mot}
\vspace{-0.2in}
\end{figure}

\textbf{Cross attention:} Inspired by ~\cite{hertz2022prompt, xie2023boxdiff}, we formulated a hypothesis: Does Flux exhibit a similar pattern where explicit cross-attentions exist between the given text prompt and intermediate attention maps within the network? As detailed in \textbf{\cref{sec:app_1}}, Flux lacks explicit cross-attention layers. Initially, this presented some challenges. However, through an in-depth examination of the neurons and features within Flux, we ultimately demonstrated (as shown in \cref{fig:mot}) that a linear relationship between text embeddings and attention maps also exists in Flux.


Specifically, as shown in \cref{eq:eq1}, the feature correlation $\mathbf{Q,K}$ is established by concatenating the textual and pixel embeddings along the last dimension: 

\begin{equation}
\begin{aligned}
& \mathbf{Q} = \mathtt{concat}(\mathbf{Q}_{text}, \mathbf{Q}_{pixel}, \mathtt{dim}=-1), \\
& \mathbf{K} = \mathtt{concat}(\mathbf{K}_{text}, \mathbf{K}_{pixel}, \mathtt{dim}=-1), \\
& \mathbf{W_{attn}} = \mathtt{Softmax}(\mathbf{Q} \times \mathbf{K}).
\end{aligned}
\label{eq:eq1}
\end{equation}

According to our experiments, we find that the nexus between text and image is inherently forged within the confines of $\mathbf{W_{attn}}$. By pinpointing the token index of the target word (want to erase) nestled within the prompt, we are capable of delineating prompt-specific characteristics. This is achieved by nullifying the pertinent column of $\mathbf{W_{attn}}$, thereby elucidating the underlying features with precision.



\begin{figure}[h]
\centering
\vspace{-0.1in}
\includegraphics[width=0.40\textwidth]{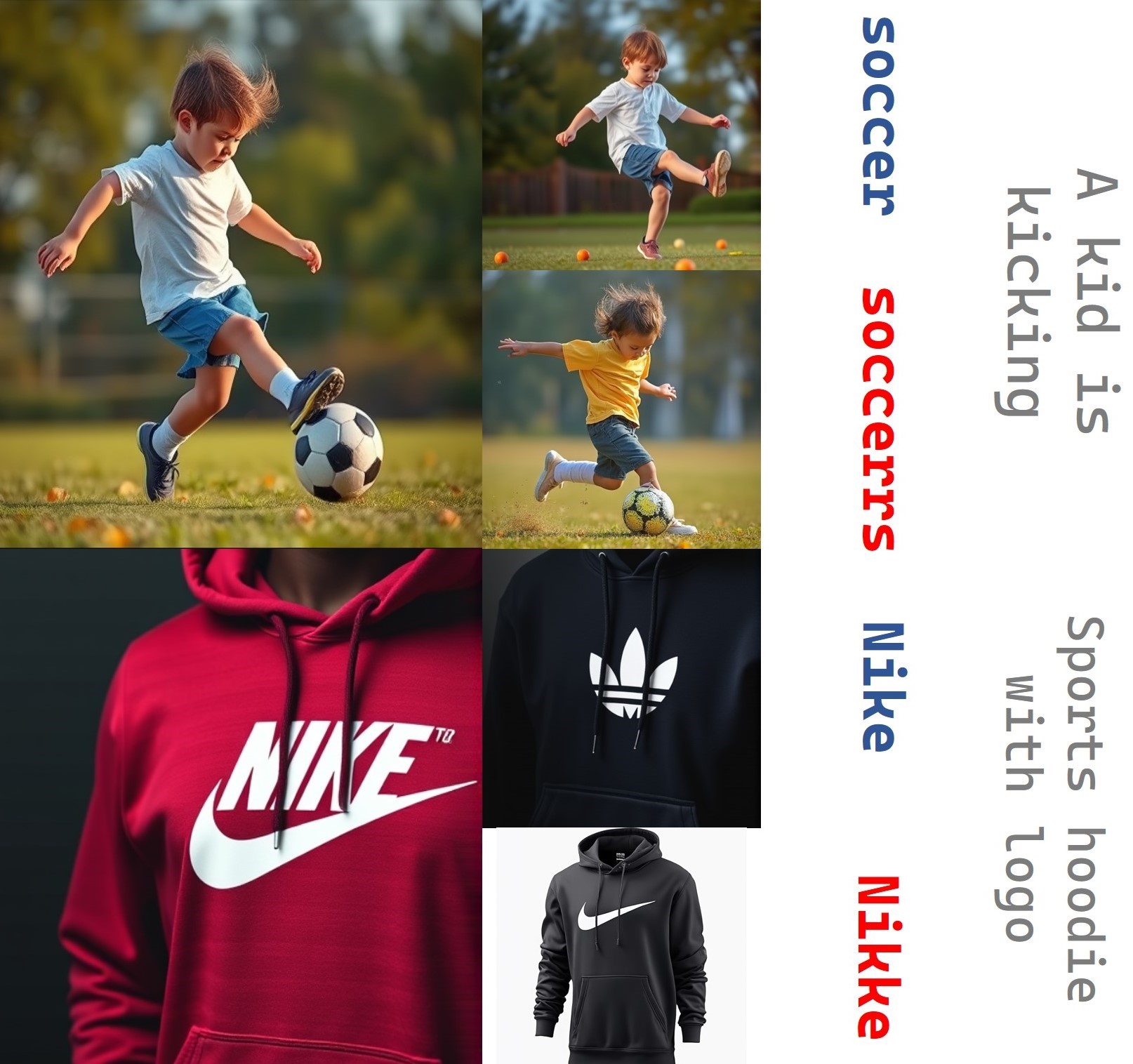}
\caption{\textbf{Attention map erasure} can be achieved by setting $\mathbf{W_{attn}}[:,:,idx_{i}] = 0, \forall i =({start}, ...,{end})$, where ${start}, {end}$ can be automatically localized given keyword \textit{e.g.} "\textcolor[rgb]{0.1, 0.2, 0.4}{\textbf{soccer}}" from input prompt "A child is kicking soccer". But this method is not generalizable when prompt is slightly modified and thus prone to be attack.}
\label{fig:mot2}
\vspace{-0.1in}
\end{figure}

\textbf{So far, Not so good:} As shown in \cref{fig:mot2}, removing a target concept seems straightforward at first: by locating the token index of the keyword in the prompt, we can delete the corresponding index column in $\mathbf{W_{attn}} \in \mathtt{[24, 1280, 1280]}$, where $\mathtt{1280 = max\_sequence\_length + head\_dim}$ and $\mathtt{24 = attn\_heads}$ (generating image resolution of $512 \times 512$). However, our experiments reveal that this technique is ineffective against one of the rudimentary prompt attack strategies: obfuscating keywords—either by altering the input prompt with nonsensical prefixes or suffixes (\textcolor[rgb]{0.1, 0.2, 0.4}{\textbf{soccer}} $\rightarrow$ \textcolor{red}{\textbf{soccerrs}}) or by introducing misspellings (\textcolor[rgb]{0.1, 0.2, 0.4}{\textbf{Nike}} $\rightarrow$ \textcolor{red}{\textbf{Nikke}}). In such cases, the erasure of the attention map proves futile, making it easy to circumvent this method and still successfully generate the target concept. (For more details, please refer to \textbf{\cref{sec:app_2}}.)


%% file: sec/method.tex
\section{Method}

\subsection{Overview}



Following our previous analysis, we have determined that the deterministic attention map erasure is vulnerable to conventional black-box attacks, rendering it less than ideal for our purposes. Consequently, we have turned our attention to a learning-based method that aims to ensure the generation quality of irrelevant concepts remains as unaffected as possible.

We address this delicate balance between removal and preservation through a bi-level optimization strategy: the lower level is designed to enhance robust concept erasure, while the upper level ensures the maintenance of irrelevant concepts. This dual-objective methodology lies at the heart of our \logopic EraseAnything.



\subsection{Bi-Level Finetuning Framework}

\textsc{Lower-Level Problem}: \textbf{Concept Erasure}

In the lower-level optimization phase, we refine the fine-tunable parameters of Flux through LoRA on the unlearned dataset $D_{un}$, which is comprised of concepts that we want to make Flux erased or unlearned.


ESD emerges as the relatively superior performer with higher negative guidance~\cite{esd}. The first sub-loss function employed in the lower-level optimization henceforth is formulated as \cref{eq:eq_esd}:

\begin{equation}
\begin{aligned}
\mathcal{L}_{esd} = \mathbb{E} \Big[ &v_{\theta_o + \Delta \theta}(x_t, c_{un}, t) \\
&- \eta \left\| v_{\theta_o}(x_t, c_{un}, t) - v_{\theta_o}(x_t, \emptyset, t) \right\|_2^2 \Big],
\end{aligned}
\label{eq:eq_esd}
\end{equation}

where $\eta$ represents the negative guidance factor, which significantly influences the degree of concept erasure. $\theta_o$ denote the parameters of the original Flux model and $\Delta \theta$ is the learnable LoRA weights for concept erasure. $x_t$ is the denoised latent code at timestep $t$ started with random noise at $x_T$ ($T$ is the total timesteps in the denoising process), $v(x_t, \emptyset, t)$ is the unconditional generation initiated with empty input prompt (\textit{a.k.a} $\emptyset = \mathtt{null \, text}$), while $c_{un} \in D_{un}$ identifies the specific concept intended for erasure, for instance, \textit{nudity}. Additionally, the term $v$ is represent the \textit{velocity} of the Flow matching process, which is the core part of Flux's scheduling mechanism and thus conceptually equivalent with the $v-prediction$~\cite{vprediction} in DMs.

Furthermore, building on the insights gleaned from the cross-attention explored in \cref{sec:sec3}, we strive to diminish the model's activations of the erased (unlearned) concepts by attenuating the attention weight allocated to keywords within the entire input prompt: $F^{un}_{idx}=\mathbf{W_{attn}}[:,:,idx]$. 

\begin{equation}
\begin{aligned}
\mathcal{L}_{attn} = \sum_{idx=start}^{end}F^{un}_{idx}.
\end{aligned}
\label{eq:eq_attn}
\end{equation}

Initially, we encountered suboptimal results because the fixed index positions of sensitive words, which we aimed to eliminate, could lead to overfitting. To counteract this, we scrambled the order of the sentences, thereby making the index positions dynamic. This method is reasonable because Flux can produce the similar content with a sentence that has been randomly shuffled. For more details, please refer to \textbf{\cref{sec:app_2}}.


\textsc{Upper-Level Problem}: \textbf{Irrelevant Concept Preservation}  

In the upper level, it serves for preserving concepts, which is fairly easy to understand: given the prompt $c$ 'a nude girl...', our objective is to eliminate the word $c_{un}$ 'nude' inside of prompt while ensuring the model can still generate an image of a unrelated concept $c_{ir}$ normally, \textit{e.g.} girl. To achieve this, we generate 6-10 images $I_{f}$ from a fixed $c$ and random seed (starting point of trajectory, same as DMs) that includes the concept to be removed (nude) and irrelevant concepts (girl), then train a LoRA (Low-Rank Adaptation) to induce shifts in the image generation process.


\begin{equation}
\begin{aligned}
\mathcal{L}_{lora} = \mathbb{E} \left[ \left\| v - v_{\theta + \Delta \theta}(u_t, c, t) \right\|_2^2 \right],
\end{aligned}
\label{eq:eq_lora}
\end{equation}

where $v = x_T - u_{pix}$, where $x_T \sim \mathcal{N}(0, I)$ and $u_{pix}$ is the VAE~\cite{vae} encoded latent code of image sampled from $I_{f}$ and $u_t = (1-t)u_{pix} + tx_T$ is the noised $u_{pix}$ at timestep $t$.

Apparently, for a broader range of irrelevant concepts, such as the abstract artistic styles and relationships mentioned earlier, this simple training recipe is insufficient to perserve the broader range of concepts that are not involved in the sentence. Considering the analysis in \cref{sec:sec3}, explicitly incorporating a collection of images and corresponding prompt lists for irrelevant concepts is cumbersome, and T5 feature is not precise enough to measure word-level similarity.

To address this, we propose a contrastive learning approach based on the attention map of keywords. This method does not require providing a set of images corresponding to irrelevant concepts. Instead, it leverages the powerful comprehension abilities of LLMs, to heuristically generate $D_{ir}$ that are irrelevant to the targeted concept for erasure. 

First, we construct a simple AI Agent that build upon on GPT-4o to sample $c_{ir} \in D_{ir}$. For efficiency reason, we then use NLTK generating the synonym of the concept that aimed to be erased, \textit{i.e.} the synonym of "nude" could be "nake". Specifically, we choose $\texttt{K}$ (default is 3) irrelevant concepts. Moving forward, we fix the sampling starting latent, i.e., $x_{T}$ as a constant value, and then substitute "nude" with "nake", $c_{ir}^{i}, i = \{1,2,3\}$ into $c$, proceeding with the denoising process independently (For more details about the $c_{ir}$ sampling, please refer to the \textbf{\cref{sec:app_3}}).

As shown in \cref{fig:mot}, we choose the attention map at higher timesteps for accurate concept-related activations. Here we get the central concept's attention feature $F^{un}$ alongside with synonym feature $F^{syn}$ and irrelevant concept set $F^{ir} = \{F^{k_1}, ..., F^{k_{K}}\}$.

Drawing inspiration from the works in ~\cite{contrastive1, contrastive2, reversion}, we have tailored the contrastive loss to function in the opposite direction, \textit{a.k.a}: \textbf{R}everse \textbf{S}elf \textbf{C}ontrastive loss (\textbf{RSC}): our training goal is to align the central feature $F^{un}$ with the dynamically shifting $F^{ir}$, while simultaneously pushing them apart from the synonym feature $F^{syn}$. The strategy here is to deviate from the conventional self-contrastive learning approach, which would typically aim to make $F^{un}$ more akin to $F^{syn}$, thereby enhancing the model's sensitivity to the term slated for removal. By inverting this approach, we aim to steer the network towards gradually discarding the concept of "nude" during learning, effectively obfuscating it within an array of irrelevant concepts.

\begin{equation}
\begin{aligned}
\mathcal{L}_{rsc} = \log\left(\frac{\sum_{i=0}^{K}\exp\left(\frac{F^{un} \cdot F^{k_i}}{\tau}\right)}{\exp\left(\frac{F^{un} \cdot F^{syn}}{\tau}\right)}\right).
\end{aligned}
\label{eq:eq_contrastive}
\end{equation}

\begin{algorithm}[h]
   \caption{BO formulation in EraseAnything}
   \label{alg:bo}
\begin{algorithmic}
   \STATE {\bfseries Input:} unlearned concept dataset and irrelevant dataset $D_{un}$ and $D_{ir}$, learning rates $\alpha_{low}, \alpha_{up}$, total iteration steps $M$.
   \FOR{$iteration=1$ {\bfseries to} $M$}
   \FOR{$c_{un}$ sampled from $D_{un}$}
   \STATE \textsc{Preparation}
   \STATE \textbf{\ding{182}} Construct a meaningful sentence $c$ involve $c_{un}$.
   \STATE \textbf{\ding{183}} Shuffle $c$ to avoid overfitting.
   \STATE \textbf{\ding{184}} Find tokenized index $idx_{start}:idx_{end}$ of $c_{un}$ from $c$.
   \STATE \textsc{Lower level: $c_{un}$ erasure}
   \STATE \textbf{\ding{185}} Update LoRA $\Delta \theta$ with \cref{eq:eq_esd}+\cref{eq:eq_attn} under $\alpha_{low}$.
   \STATE \textsc{Upper level: $c_{ir}$ preserving}
   \STATE \textbf{\ding{186}} Retrieve $c_{ir}$, $c_{syn}$ \textit{w.r.t} to $c_{un}$ and replace them into $c$ separately to have $F^{ir,syn}$.
   \STATE \textbf{\ding{187}} Update LoRA $\Delta \theta$ with \cref{eq:eq_lora}+\cref{eq:eq_contrastive} under $\alpha_{up}$.
   \ENDFOR
   \ENDFOR
\end{algorithmic}
\end{algorithm} 

As depicted in \cref{eq:eq_contrastive} (detailed derivations are provided in \textbf{\cref{sec:app_4}}), $\tau$ is the temperature hyperparameter that governs the model's capacity to differentiate between irrelevant concepts. A high $\tau$ causes the contrastive loss to treat all irrelevant concepts with equal importance, potentially resulting in a lack of focus in the model's learning process. Conversely, a low $\tau$ may cause the model to concentrate excessively on especially challenging irrelevant concepts, which could be mistaken for potential synonym sample. Based on empirical testing, we have determined that setting $\tau=0.07$ is optimal for our model's performance.

\textbf{Bi-Level Optimization:} As shown in \cref{eq:eq_contrastive}, the last loss term defined in our method is finalized. Integrating the
aforementioned two optimization problems, we have a bi-level
optimization illustrated in \cref{eq:eq_bilevel} (please check \cref{alg:bo} for details).

\begin{equation}
\begin{aligned}
\min \mathcal{L}_{lora+rsc}(\Delta^{*} \theta; D_{ir}) &
 \\ \textit{s.t.} \quad 
\Delta^{*} \theta = \min \mathcal{L}_{esd+attn}(\Delta \theta; &D_{un})
\end{aligned}
\label{eq:eq_bilevel}
\end{equation}









%% file: sec/experiment.tex
\section{Experiments}

Here, we conduct a comprehensive evaluation of EraseAnything, benchmarking it on various tasks, ranging from concrete to abstract: \textit{e.g.}, soccer, architecture, car to artistic style, relationships and \textit{etc}.



\subsection{Implementation Details}

We have opted for the Flux.1 [dev] model with publicly accessible network architecture and model weights, a distilled version of Flux.1 [pro] that retains high quality and strong prompt adherence. Our codebase utilizes widely adopted diffusers~\cite{diffusers}, a popular choice among developers and researchers for DMs. Unless otherwise specified, our experiments employ the flow-matching Euler sampler with 28 steps and AdamW~\cite{adamw} optimizer for 1,000 steps, with a learning rate $\alpha_{low} =0.001, \alpha_{up}=0.0005$ and an erasing guidance factor $\eta=1$ under all conditions. 

In terms of concept construction, we harness the power of NLTK~\cite{nltk} to generate synonym concepts, and we employ GPT-4o in the extraction of irrelevant concepts. Our fine-tuning process focuses on the text-related parameters \texttt{add\_q\_proj} and \texttt{add\_k\_proj} (subsets of $\mathbf{Q}$ and $\mathbf{K}$) within the dual stream blocks. Furthermore, EraseAnything requires minimal learnable weights compared to methods such as ESD, with only \textbf{3.57MB} allocated per concept. The model is trained on NVIDIA A100 (80GB VRAM) GPU with batch size 1.




\begin{figure}[t]
\centering
\includegraphics[width=0.36\textwidth]{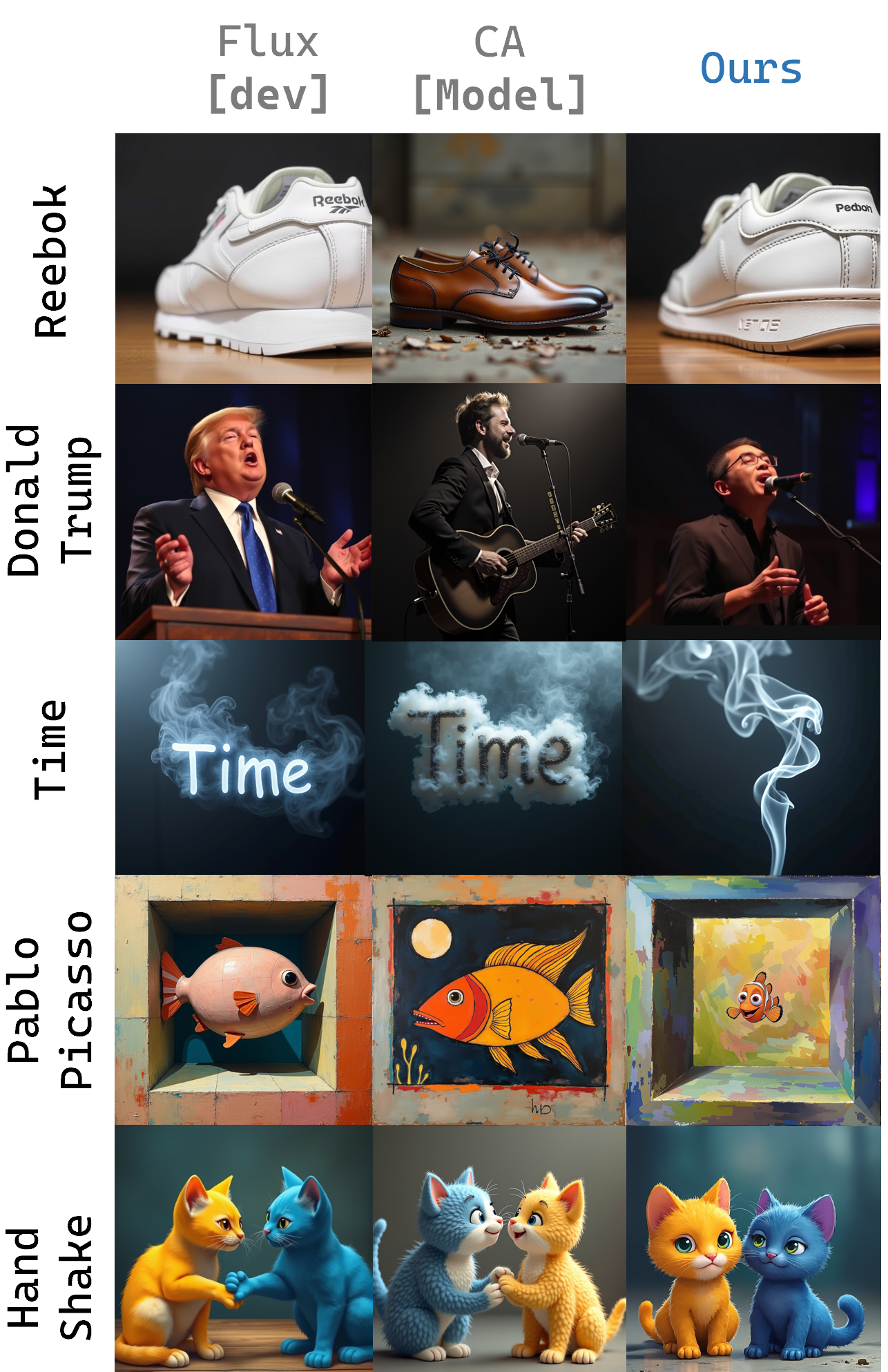}
\caption{\textbf{Single-concept erasure}. We test our model across three levels of granularity—Entity, Abstraction, and Relationship—to assess its effectiveness. Furthermore, we have incorporated the versatile CA~\cite{ca} [model] to enhance the visual contrast for a clearer comparison.}
\label{fig:exp_vis}
\vspace{-0.2in}
\end{figure}


\begin{figure}[t]
\centering
\includegraphics[width=0.44\textwidth]{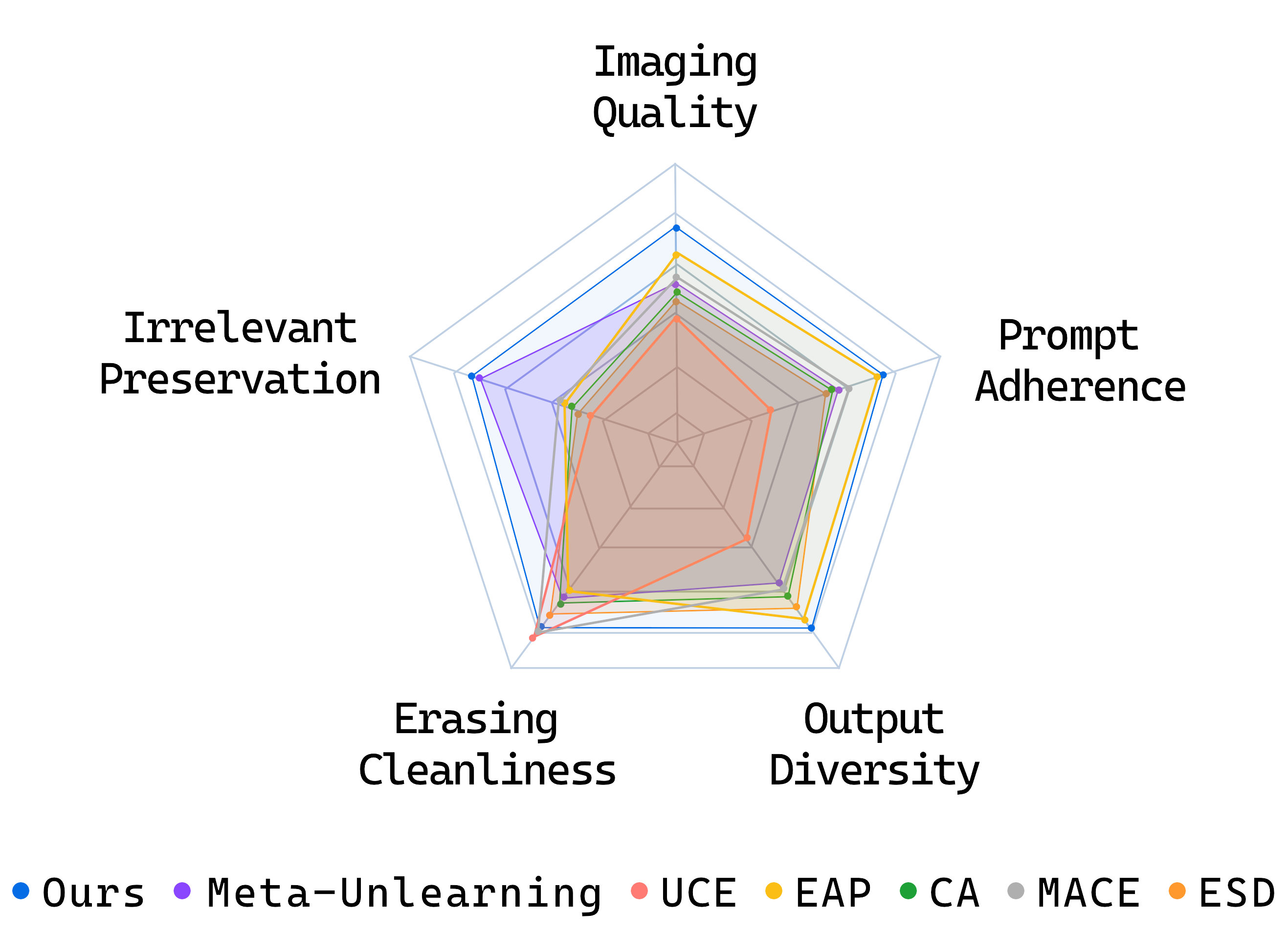}
\caption{\textbf{User Study}. We have created an interface (see \textbf{\cref{sec:app_5_us}} for details) that shows the users with AIGC contents under various methods that transplanted to Flux. With a scoring system where 1 (worst) and 5 (best), it is clear that EraseAnything offers the best overall performance when assessed across five different dimensions.}
\label{fig:user_study}
\end{figure}

\begin{table*}[t]
\caption{Assessment of Nudity Removal: (Left) Quantity of explicit content detected using the NudeNet detector on the I2P
benchmark. (Right) Comparison of FID and CLIP on MS-COCO. The performance of the original Flux [dev] is presented for reference.}
\label{sample-table}
\vskip 0.15in
\begin{center}
\begin{small}
\begin{sc}
\begin{tabular}{lccccccr}
\toprule
\multirow{2}{*}{\centering Method} & \multicolumn{4}{c}{Detected Nudity (Quantity)} & \multicolumn{2}{c}{MS-COCO 10K} \\
& Common & Female & Male & Total$\downarrow$ & FID$\downarrow$ & CLIP$\uparrow$ \\
\midrule
CA (Model-based)~\cite{ca}   & 253 &65 & 26 & 344 & 22.66 & 29.05 \\
CA (Noise-based)~\cite{ca}  & 290 & 72 & 28 & 390 & 23.07 & 28.73\\
ESD~\cite{esd}    & 329 & 145 & 32 & 506 & 23.08 & 28.44\\
UCE~\cite{uce}    & 122 & 39 & 12 & 173 & 30.71 & 24.56\\
MACE~\cite{lu2024mace} & 173 & 55 & 28 & 256 & 24.15 & 29.52\\
EAP~\cite{eap}    & 287 & 86 & 13 & 386 &   22.30   & 29.86\\
Meta-Unlearning~\cite{gao2024meta} & 355 & 140 & 26 & 521 & 22.69 & 29.91\\
\rowcolor{blue!10}
Ours      & 129 & 48 & 22 & 199  &  21.75    & 30.24\\
\midrule
Flux.1 [dev]   & 406 & 161 & 38 & 605 & 21.32 & 30.87 \\
\bottomrule
\end{tabular}
\end{sc}
\end{small}
\end{center}
\vskip -0.1in
\label{tab:nudenet_detection}
\end{table*}

\subsection{Results}

\textbf{Nudity Erasure} serves as a well-established benchmark that has gained widespread recognition. To assess the effectiveness and versatility of our approach, we begin by applying it to the task of nudity erasure. Specifically, we used our concept-erased model to generate images from a comprehensive set of 4,703 prompts extracted from the Inappropriate Image Prompt (I2P) dataset~\cite{sld}. For the identification of explicit content within these images, we deploy NudeNet~\cite{bedapudi2019nudenet}, using a detection threshold of \textbf{0.6}. Furthermore, to evaluate the specificity of our method in regular content, we randomly select 10,000 captions from the MS-COCO captioning dataset (validation)~\cite{mscoco}. Finally, we generate images from these captions and assess the results using both the Fréchet Inception Distance (FID) and CLIP scores.

\cref{tab:nudenet_detection} presents our results in comparison with the current state-of-the-art algorithms. It is evident that our method generates the second-lowest amount of explicit content when conditioned on 4,703 prompts, only outperformed by the UCE. Yet, it stands out with remarkable FID and CLIP scores, suggesting that our approach exerts a minimal negative influence on the original model's ability to generate regular content. In contrast, the UCE, while leading in explicit content reduction, shows a sharp decline in efficacy according to these metrics.



\begin{figure}[hpbt]
\centering
\includegraphics[width=0.5\textwidth]{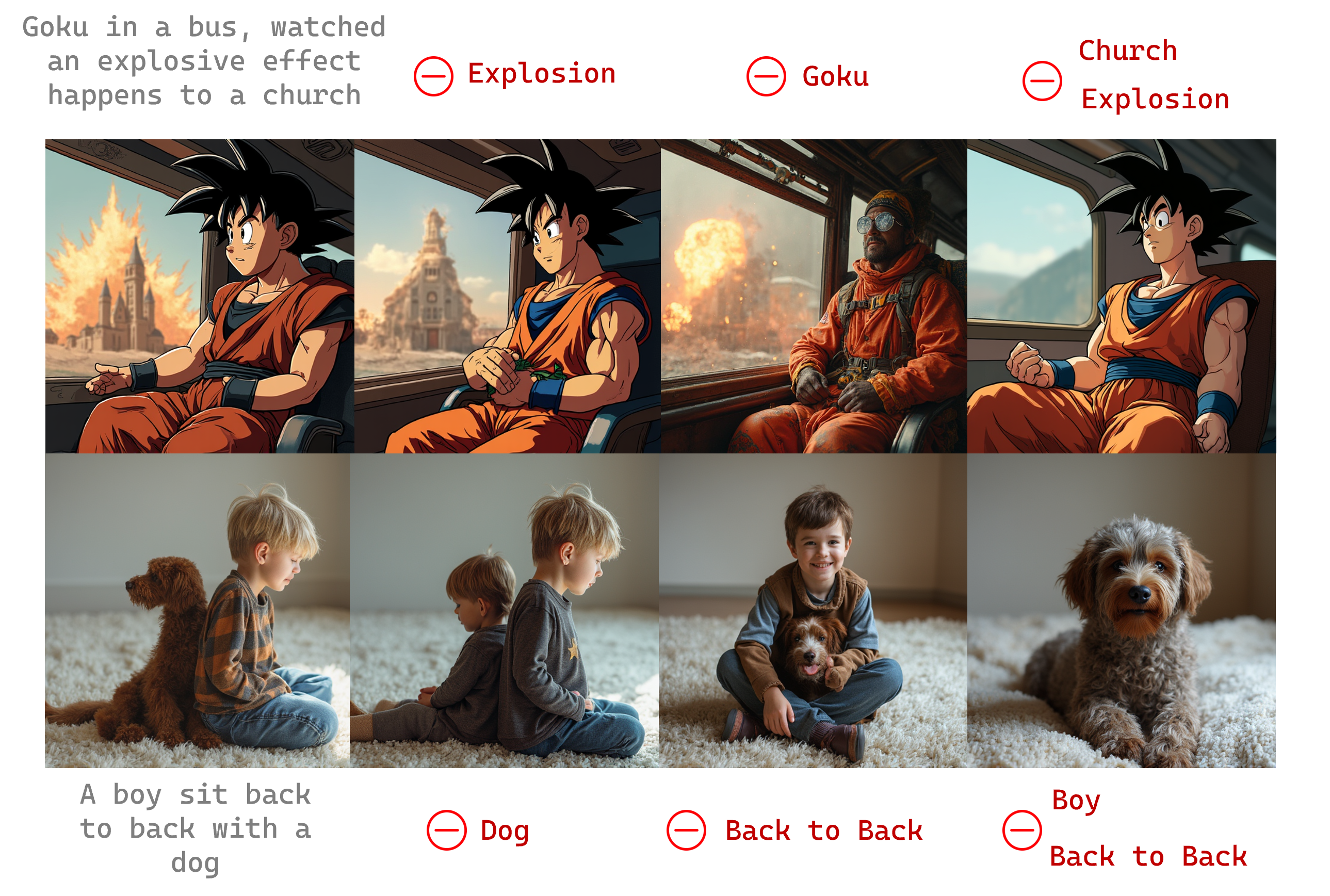}
\caption{\textbf{Multi-concept erasure}.}
\label{fig:exp_multi}
\vspace{-0.1in}
\end{figure}

\textbf{Miscellaneousness Erasure} In this section, we evaluate our method on 3 conceptual categories: \textbf{Entity}, \textbf{Abstraction} and \textbf{Relationship}. Here, we choose 10 concept for each category (Please check \textbf{Appendix C} for the full list of concepts) and adopt the measuring metrics described in \cref{tab:table_mis}. As shown in \cref{fig:exp_vis} and \cref{fig:exp_multi}, our method can effectively remove a variety of concepts (including multiple-concepts!) while maintaining minor disturbance compared to CA, which substantiates the claim: \textbf{EraseAnything is truly an "Erase Anything" solution}.

The findings presented in \cref{tab:table_mis} reveal that our method outperforms the traditional CA in terms of erasure efficacy, the retention of unrelated concepts, and the robustness against synonym substitution. This underscores the ability of our method to not only grasp the targeted concepts for erasure but also to discern those that are semantically adjacent, all while exerting an imperceptible negative influence on the model's intrinsic capabilities. For a comprehensive evaluation of our model's robustness, kindly refer to \cref{sec:app_2}.



\textbf{User Study} To gauge the human perception of the effectiveness of our method, we conducted a user study with five dimensions, where each focusing on a different aspect of erased model. For the first two trials: Erasing Cleanliness (prompt with $c_{un}$ and generated images do not contain concept around $c_{un}$) and Irrelevant Preservation (prompt with $c_{ir}$ can be normally generated), we utilized the same concepts categorized under Entity, Abstraction, and Relationship. For each concept, images were generated using the same random seed across all methods, ensuring a fair comparison.

Our study involved 20 non-artist participants, each providing an average of 200 responses. \cref{fig:user_study} shows that our method exhibited a comprehensive performance, achieving outstanding results across all 5 aspects, thus making EraseAnything a good all-round player in concept erasure area.

As for the settings, please refer to the \cref{sec:app_5_us} for detailed information about it due to the page limit. 





\begin{table}[hb]
\caption{Evaluation of Erasing the specific category: Entity (\textit{e.g.} soccer), Abstraction (\textit{e.g.} artistic style) and Relationship (\textit{e.g.} kiss) are presented. CLIP classification accuracies are reported for each erased category in three sets: the erased category itself (Acc$_e$, efficacy), the remaining unaffected categories (Acc$_{ir}$, specificity) and synonyms of the erased class (Acc$_g$, generality). All presented values are denoted in percentage (\%).}
\vskip 0.15in
\begin{center}
\begin{small}
\begin{sc}
\begin{tabular}{lccr}
\toprule
Method & Acc$_{e}$ $\downarrow$ & Acc$_{ir}$ $\uparrow$ & Acc$_{g}$ $\downarrow$ \\
\midrule
CA (Entity)    & 14.8 & 89.2 & 27.3 \\
CA (Abstraction)    & 25.2 & 88.3 & 29.6 \\
CA (Relationship)    & 22.7 & 88.6 & 23.1 \\
\midrule
\rowcolor{blue!10}
Ours (Entity)      & 12.5 & 91.7 & 18.6  \\
\rowcolor{blue!10}
Ours (Abstraction)      & 21.1 & 90.5 & 24.7  \\
\rowcolor{blue!10}
Ours (Relationship)      & 18.4 & 90.2 & 19.3  \\
\bottomrule
\end{tabular}
\end{sc}
\end{small}
\end{center}
\vskip -0.1in
\label{tab:table_mis}
\end{table}

\subsection{Ablation study}


To assess our loss functions, we conducted an ablation study on the task of celebrity image erasure. We chose a subset from the CelebA~\cite{celeba}, omitting those that Flux [dev] couldn't accurately reconstruct. This resulted in a dataset of 100 celebrities, split into two groups: 50 for erasure and 50 for retention. Unlike MACE's massive concept erasure, EraseAnything is trained on individual celebrities. Performance was evaluated by averaging metrics from \cref{tab:table_ablation}. 


Different variations and their results are presented in \cref{tab:table_ablation}. $\mathcal{L}_{esd}$ itself fall short of the complete erasure of target concept, resulting in a not so low ACC$_{e}$. With the addition of $\mathcal{L}_{attn}$, ACC$_{e}$ has fallen dramatically but the retention of irrelevant concepts was fail \textit{w.r.t} ACC$_{ir}$. Incorporating the loss term $\mathcal{L}_{rsc}$, we introduce a approach that may lead to achieving high ACC$_{ir}$ values. By organically combining all these loss terms, we achieve a comprehensive model that consistently demonstrates the lowest ACC$_{e}$ and the highest ACC$_{ir}$ compared to previous configurations.


\textbf{Others}. Due to the page limits, we put remaining experimental details and results in \textbf{\cref{sec:app_6}}. This includes the visualizations under different configs; the complete list of celebrities used in ablation study and a full set of visualizations upon conceptions from various subjects.

\begin{table}[hbpt]
\caption{Ablation Study on Erasing Celebrities, we ablate four loss terms used in our experiments. A celebrity recognition is trained to measure the accuracies \textit{w.r.t} the erased celebrity (Acc$_e$, efficacy) and the remaining unaffected celebrities (Acc$_{ir}$, specificity). All
presented values are denoted in percentage (\%).}
\begin{center}
\begin{small}
\begin{sc}
\begin{tabular}{lccr}
\toprule
Config & Acc$_{e}$ $\downarrow$ & Acc$_{ir}$ $\uparrow$ \\
\midrule
$\mathcal{L}_{esd}$ + $\mathcal{L}_{attn}$    & 15.3 & 82.1  \\
$\mathcal{L}_{esd}$ + $\mathcal{L}_{lora}$    & 20.5 & 77.9  \\
$\mathcal{L}_{esd}$ + $\mathcal{L}_{rsc}$    & 16.1 & 85.6  \\
$\mathcal{L}_{attn}$ + $\mathcal{L}_{rsc}$    & 18.6 & 81.7  \\
$\mathcal{L}_{attn}$ + $\mathcal{L}_{lora}$ + $\mathcal{L}_{rsc}$   & 15.8 & 80.2  \\
\midrule
\rowcolor{blue!10}
Full      & \textbf{14.9} & \textbf{88.5}  \\
\bottomrule
\end{tabular}
\end{sc}
\end{small}
\end{center}
\vskip -0.1in
\label{tab:table_ablation}
\end{table}

\section{Limitations}

Although "EraseAnything" has demonstrated its formidable ability to erase concepts across various domains, we have identified challenges it faces in certain situations:

\textbf{Extensive Concept Erasure}: When tasked with erasing multiple concepts simultaneously, such as 10 or more concepts (LoRAs), the \textbf{Normalized Sum} strategy, as depicted in \cref{eq:lora_add}, results in a proportional decrease in the impact of each concept's erasure. Consequently, a significant and important avenue for research in this field is to explore efficient methods for combining a large number of LoRAs (more than 100) effectively.

\textbf{Fine-grained Control}: Another issue pertains to the inability to guarantee the strength of the erasure during fine-tuning. This is an uncharted yet intriguing area in the realm of concept erasure, which could provide us with a deeper understanding of the concept formulation. It would also enable more precise control over the erasure process, \textit{e.g.} a slider could be provided to control the intensity during interactive concept erasure.

%% file: sec/conclusion.tex
\section{Conclusion}

In this paper, we propose \logopic \textbf{EraseAnything}, a Flux-based concept erasing method. Leveraging a bi-level optimization strategy,
we strike a balance between erasing the target concept that bound to be removed while preserving the irrelevant concepts unaffected, mitigating long-lasting notorious risk of overfitting and catastrophic forgetting. Experiments across diverse tasks strongly demonstrate the effectiveness and versatility of our method.

\section{Acknowledgement}

We would like to express our sincere gratitude to Xingchao Liu from the University of Texas at Austin for his invaluable contributions and unwavering support throughout our research endeavor. His expertise, particularly in the field of rectified flow, has been instrumental in helping us navigate and avoid potential pitfalls. Furthermore, our thanks go to Eliza (ai16z)~\footnote{https://github.com/ai16z/eliza}, an AI Agent framework that has been integral to our study. Specifically, we have employed Eliza to generate our charming icon \logopic!.